\documentclass[runningheads]{llncs}

\usepackage{eccv}
\usepackage{eccvabbrv}

\usepackage{graphicx}
\usepackage{booktabs}
\usepackage{colortbl}
\usepackage{tabularx}
\usepackage{multirow}
\usepackage[utf8]{inputenc}
\usepackage{booktabs}

\usepackage[accsupp]{axessibility}  % Improves PDF readability for those with disabilities.

\usepackage[hidelinks,colorlinks,linkcolor=eccvblue,urlcolor=eccvblue,citecolor=eccvblue,anchorcolor=eccvblue]{hyperref}

\usepackage{orcidlink}

\begin{document}

\newcommand{\NAME}{{ZED}}
\newcommand{\OURS}{\NAME}

% ---------------------------------------------------------------

\title{Zero-Shot Detection of AI-Generated Images}

\titlerunning{Zero-Shot Detection of AI-Generated Images}

\author{
Davide Cozzolino\inst{1} \and 
Giovanni Poggi\inst{1} \and
Matthias Nie\ss ner\inst{2} \and \\ 
Luisa Verdoliva\inst{1,2}
}

\authorrunning{Cozzolino et al.}

\institute{University Federico II of Naples, 80125 Naples, Italy \and Technical University of Munich, 85748 Garching, Germany \\
\email{\{davide.cozzolino,poggi,verdoliv\}@unina.it}, \email{\ niessner@tum.de} }

\newcommand{\ru}{\rule{0mm}{3mm}}
\newcommand{\rota}[1]{\rotatebox[origin=c]{90}{#1}}
\newcommand{\band}{\rowcolor{gray!15}}
\newcommand{\graycell}{\cellcolor{gray!15}}
\definecolor{MyDarkGray}{rgb}{0.5,0.5,0.5}
\definecolor{tabvline}{HTML}{a8a495}
\definecolor{prompt_blue}{HTML}{1f78b4}
\definecolor{prompt_red}{HTML}{d45c43}
\renewcommand{\b}[1]{\textbf{#1}}
\renewcommand{\u}[1]{\underline{#1}}
\newcommand{\NLL}{{\rm NLL}}
\newcolumntype{C}[1]{>{\centering\arraybackslash}p{#1}}
\newcolumntype{L}[1]{>{\raggedleft\arraybackslash}p{#1}}

\maketitle

\begin{abstract}
Detecting AI-generated images has become an extraordinarily difficult challenge as new generative architectures emerge on a daily basis with more and more capabilities and unprecedented realism. 
New versions of many commercial tools, such as DALL·E, Midjourney, and Stable Diffusion, have been released recently, and it is impractical to continually update and retrain supervised forensic detectors to handle such a large variety of models. 
To address this challenge, we propose a zero-shot entropy-based detector (\NAME) that neither needs AI-generated training data nor relies on knowledge of generative architectures to artificially synthesize their artifacts.
Inspired by recent works on machine-generated text detection, our idea is to measure how {\em surprising} the image under analysis is compared to a model of real images. 
To this end, we rely on a lossless image encoder that estimates the probability distribution of each pixel given its context. 
To ensure computational efficiency, the encoder has a multi-resolution architecture and contexts comprise mostly pixels of the lower-resolution version of the image.
Since only real images are needed to learn the model, the detector is independent of generator architectures and synthetic training data.
Using a single discriminative feature, the proposed detector achieves state-of-the-art performance. 
On a wide variety of generative models it achieves an average improvement of more than 3\% over the SoTA in terms of accuracy. Code is available at \url{https://grip-unina.github.io/ZED/}.

\end{abstract}

\section{Introduction}
\label{sec:intro}

The quality of AI-generated images has improved tremendously in recent years, to the point where they are virtually indistinguishable from real images upon visual inspection.
In addition, the latest generators are widely available online and allow easy creation and retouching of images based on simple textual prompts.
All this opens the way to endless application opportunities in a variety of fields, from the creative arts to industries of all kinds.
However, on the flip side, such tools can be also used for malicious purposes, thus posing serious threats to our society.
For example, pre-trained generators can be easily optimized to generate fake works by a specific artist ~\cite{heikkila2022this}, or used to orchestrate effective, large-scale disinformation campaigns to influence public opinion in advanced democracies~\cite{epstein2023art}.
These immediate risks create an urgent need for reliable and automated detection of AI-generated images~\cite{lin2024detecting}.

Until very recently, supervised learning paradigms dominated the image forensics community, with deep models trained on large datasets of real and fake images \cite{roessler2019ff++}. 
These approaches, however, are tailored to specific domains and are difficult to generalize to unseen deepfake samples.
In the seminal paper by Wang \etal \cite{wang2020cnn}, it is shown that a simple detector trained only on ProGAN images from 20 different categories generalizes well to other images created by different generative adversarial networks (GAN) thanks to suitable augmentation.
However, performance still suffers on images generated by prompt-driven diffusion models (DM). 
Similarly, a detector suitably trained on Latent DM images performs well on all other DM images but fails to generalize properly on GAN images \cite{corvi2023detection}.
To reduce the dependence on training data, recent works \cite{amoroso2023parents, ojha2023towards, sha2022fake, cozzolino2023raising} rely on general-purpose features extracted by pre-trained visual-language models, such as CLIP (Contrastive Language-Image Pre-Training) \cite{radford2021learning}. Despite the good performance, these methods still depend on the choice of the training dataset.
A recent trend to improve generalization is based on few-shot methods \cite{cozzolino2018forensictransfer, du2020towards, jeon2020tgd} which can partially solve the problem, but still require some prior knowledge of the target models, even if limited to a few images. 
With this work we make a step further and develop an approach that is not influenced at all by newer and previously unseen generative models.

\begin{figure}[t!]
    \centering
    \includegraphics[page=1, height=0.26\linewidth, trim=185 320 185 0,clip]{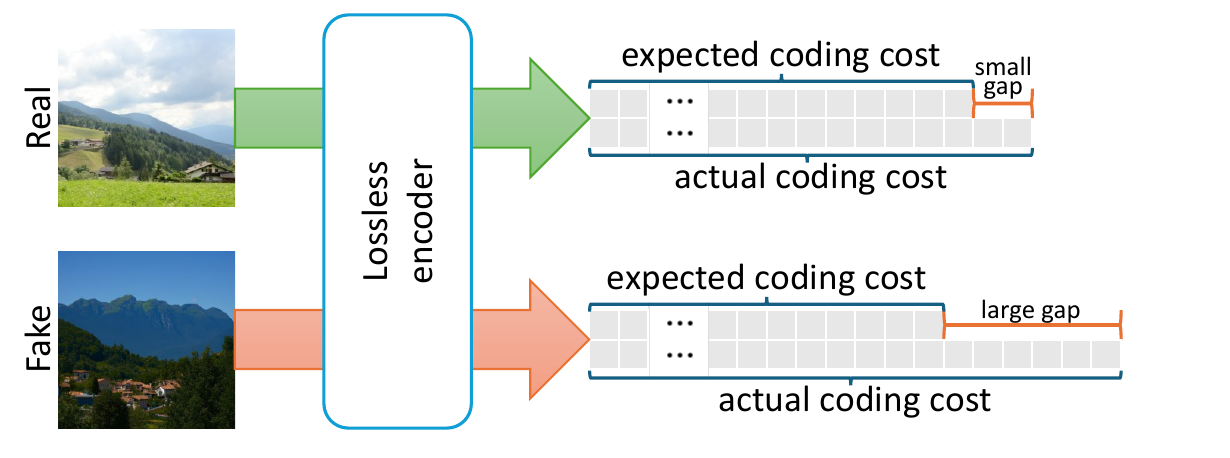}
    \includegraphics[page=1, height=0.26\linewidth, trim=0 0 0 0,clip]{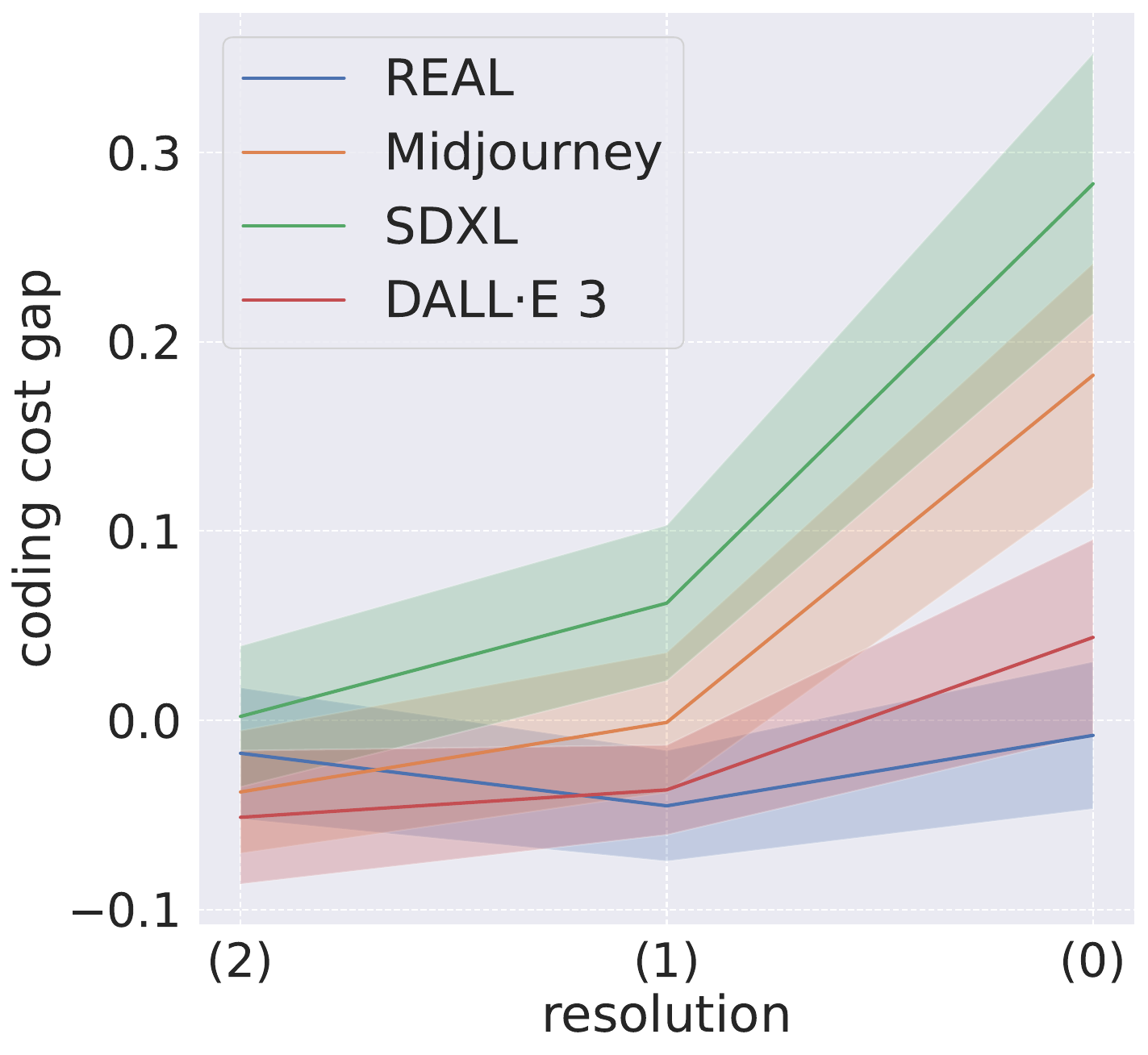}
    \hfill
\caption{
\NAME\ leverages the intrinsic model of real images learned by a state-of-the-art lossless image coder. 
For real images, the model is correct and the actual coding cost is close its expected value.
Synthetic images have different statistics than real images, so they ``surprise'' the encoder, and the actual coding cost differs significantly from its expected vale. 
This is evident from the graphic on the right that shows how the coding cost gap increases for synthetic images much more than for real ones when predicting high resolution details from low resolution data.
}
\label{fig:teaser}
\vspace{-0.5cm}
\end{figure}

To this end, we propose a zero-shot detection method that only requires real images for learning their underlying distribution.
Our key idea is to use lossless coding and a multi-resolution prediction strategy for computing conditional distributions of all image pixels at three different levels of resolution. 
Given such distributions, we compute statistics related to the actual and expected coding cost. 
If the image is coherent with the predicted distribution ({\em no surprise}), then there is no mismatch and the image under analysis is labelled as real. 
We expect synthetic images to be characterized by a higher coding cost under the distribution of real images (see Fig.\ref{fig:teaser}). 
Based on this intuition, we design discriminative features that measure how well the image under test fits the model of real images embedded in the encoder.
Even by using a single feature, we can obtain significant performance above 95\% in terms of AUC for several recent models, such as DALL·E, Midjourney, and SDXL.

In summary, the main contributions of this paper are the following:
\begin{itemize}
\item  we propose a zero-shot detector of artificially generated images:
no fake images are necessary for training which guarantees independence from any specific generation method;
\item  this is the first work that exploits an implicit model of real images, learnt for lossless encoding to address image forensics task;
\item  our experiments show on a wide variety of generative models that even using a single feature the proposed detector provides state-of-the-art results (+3.4\% in terms of accuracy).
\end{itemize}

\section{Related work}
\label{sec:related_work}

\vspace{-3mm}
\subsubsection{Supervised learning.} The problem of distinguishing synthetic images from real ones is commonly formulated as a binary classification task. State-of-the-art methods explicitly or implicitly exploit forensic artifacts by leveraging a large amount of real and generated images. Some of them rely on semantic flaws, such as face asymmetries \cite{bohacek2023geometric} or incorrect perspective, lighting, shadows \cite{farid2022lighting, farid2022perspective, sarkar2023shadows}.
However, technology advances very quickly and such errors will very likely disappear in next-generation tools.
Therefore, most methods focus on low-level and inconspicuous artifacts \cite{durall2020watch, corvi2023intriguing}.
Major efforts have been made to prevent conventional supervised detectors from overfitting the training data.
Popular recipes include using datasets as varied as possible with intense augmentation \cite{wang2020cnn}, pre-training models on large general-purpose datasets \cite{mandelli2022detecting}, 
preserving fine-grain details of images \cite{gragnaniello2021GAN, chai2020what},
exploiting high-frequency artifacts in the spatial \cite{tan2023learning, liu2022detecting, sinitsa2023deep} or Fourier domain \cite{zhang2019detecting, frank2020leveraging, durall2020watch}, leveraging inter-pixel correlation discrepancies \cite{zhong2023rich, tan2023rethinking}, adopting inversion techniques \cite{albright2019source, wang2023dire}.

With the advent of diffusion models that presents significant architectural differences with GANs, the importance to design methods that work equally well on known and unknown sources became even more evident \cite{corvi2023detection}. An important finding was the increased generalization that could be achieved using pre-trained large vision-language models, such as CLIP-ViT \cite{ojha2023towards}.
In this case only a lightweight linear classifier is trained on top of these features to adapt to the forensic task.
Very good performance is obtained on DMs even if the network was trained only on GANs. Other methods also show the potential of such approach \cite{ricker2022towards, amoroso2023parents, cozzolino2023raising}, sometimes including multimodal features \cite{sha2022fake, liu2023forgery}. 

Some supervised methods assume to have only real images available and create the synthetic images needed for training by simulating the artifacts introduced by a generator, for example by passing real images through an autoencoder \cite{zhang2019detecting, frank2020leveraging, jeong2022FingerprintNet}. The more generative architectures are simulated, the more effective is the detector. Of course, the performance degrades on images generated by an architecture not considered in the simulation phase. Differently from all these methods our approach does not require collecting or generating synthetic images thus avoiding any type of dependence on this class.

\vspace{-3mm}
\subsubsection{Few-shot/incremental learning.} 
A significant step towards improved generalization is the use of few-shot or incremental learning strategies \cite{cozzolino2018forensictransfer, marra2019incremental, du2020towards, jeon2020tgd}. 
Along this path, a recent work \cite{epstein2023online} proposes to regularly re-train a detector on new synthetic generators in the very same temporal order of their release, as in a real-world scenario.
Results show a good generalization to unseen models, but only as long as the architecture of new generators is similar to that of old ones.
Although few-shot methods represent an important progress in reducing the dependence on training data, the ultimate goal is to remove this dependence entirely to ensure maximum generalization. In pursuit of this goal, in this work we propose a truly zero-shot detector.

\vspace{-3mm}
\subsubsection{Zero-shot learning.}
Only a few very recent papers avoid training on synthetic data altogether.
A solution was proposed in \cite{ricker2024aeroblade} based on the observation that synthetic images are reconstructed more accurately than real images by a latent DM autoencoder. The main limitation is that the method only reliably detects images generated by latent diffusion models.
The method in \cite{he2024rigid}, instead, exploits the fact that small perturbations of [real/synthetic] images correspond to [small/large] variations in the embedding space of a pre-trained large model.
Differently from these strategies our work takes inspiration from some interesting proposals that have recently appeared for synthetic text detection \cite{solaiman2019release, gehrmann2019gltr, mitchell2023detectGLPT, hans2024spotting}.
They exploit the fact that LLMs (Large Language Models) work by generating the probability distribution of the next token given the previous ones.
In the generation phase, new tokens are sequentially added to a sentence based on these distributions.
In the analysis phase, one can replicate the process for a given sentence under test and measure how well the actual tokens match the predicted ones.
A good match suggests that the sentence was indeed generated by an LLM.
Although inspired by these methods,
our zero-shot synthetic image detector differs from them because it leverages a model of {\em real} images and does not depend in any way on synthetic data or generators. 
Moreover, to build the model we take advantage of the remarkable field-proved ability of lossless encoders to accurately describe pixels based on their context.

\section{Method}
\label{sec:method}

\subsection{Background}

\newcommand{\x}{X}
Here we provide some background on zero-shot methods that leverage large pre-trained language models for machine-generated text detection.
They exploit the native functionality of these models to provide next-token predictions \cite{hans2024spotting}.
Before a string of characters $s$ can be processed by a language model, it must be parsed into a sequence of tokens (mostly words).
The tokenizer $T$ outputs a list of indices
\begin{equation}
    T: s \to \{x_0, x_1,\ldots, x_L\},
\end{equation}
where $x_i \in \{1,...,n\}$ is the index of the $i$-th token of the sequence, addressing a size-$n$ vocabulary of tokens.
The language model operates by predicting the next index/token given the list of previous ones, thereby allowing for the generation of a full sentence given just a short prompt.
Actually, language models output more information than just the index of the most likely token.
Given the list of previous indices $X_i=\{x_0,\ldots,x_{i-1}\}$, 
they provide the probability of all possible values of the current one, that is, $P(x_i=k | X_i)$, for $k=1,\ldots,n$.

The idea is to exploit this functionality to measure the conformity of the string under analysis to the LLM intrinsic model of language.
That is, these methods try to answer the question ``How likely is it that this sentence was generated by my LLM?''
Hence they compute (for free) the likelihood of the given list of indices under the probability distribution learned by the LLM
\begin{equation}
    P(x_0,\ldots,x_L) = P(x_0)\cdot P(x_1|x_0)\cdot \ldots \cdot P(x_L|x_0,\ldots,x_{L-1}) = P(x_0)\prod_{i=1}^L P(x_i|X_i)
\end{equation}
In practice, the negative log-likelihood (also called log-perplexity) is computed instead, that is (neglecting $x_0$)
\begin{equation}
    {\rm NLL} = -\sum_{i=1}^L \log P(x_i|X_i)
\end{equation}
If the $i$-th observed index $x_i$ was very likely to come after the previous ones, namely, it is not surprising, its contribution to the NLL is close to 0.
On the contrary, if it was unlikely to appear, given the previous ones (an anomaly) it impacts significantly on the NLL.
Overall, a sequence with low NLL is likely to have been generated by the LLM, and will be therefore detected as synthetic.
Of course, this basic description is only meant to convey the general concepts,
the reader is referred to the literature \cite{ghosal2023towards} for more details. 

\subsection{From Text to Images}

When we try to translate the above concepts into the realm of images, we run into a big problem: the most effective and popular image generation engines {\em do not} provide anything similar to the next token distribution observed in the case of LLMs.
Indeed, there exist some autoregressive synthesis methods \cite{reed2017parallel, mahajan2021pixelpyramids} that could be adapted to this task, but their generation approach is very different from those of the most popular GAN- and DM-based methods.
Therefore in this work we change perspective or, better said, we now assume the correct one-class perspective, and look for a model of real images, rather than synthetic ones.
Armed with such a model, we will be able to decide whether a given image is unsurprising, therefore real, or somewhat anomalous, therefore synthetic, regardless of the specific generation model used to create it.

Now, the concepts of prediction, surprise, perplexity, along with information measure and entropy, are pervasive in the literature on image coding, part of information theory.
Lossless image encoders typically include a predictor that, given a suitable context, estimates the value of the target pixel, and an entropy encoder that efficiently represents prediction errors.
Indeed, by analyzing the recent literature in the field we managed to single out a tool that perfectly suits our needs,
the Super-Resolution based lossless Compressor (SReC) proposed by Cao \etal \cite{cao2020lossless},
which provides a computationally lightweight tool for predicting the distribution of image pixels at multiple resolution.

\subsection{Super-resolution based Lossless Compressor}

Here we present a high-level description of SReC, focusing only on the aspects more relevant for our purposes.
The interested reader is referred to the original paper for details \cite{cao2020lossless}.
The general idea is to train a neural network to predict the current pixel, $x_{i,j}$, given a set of previously coded pixels,
and encode the difference between the true pixel value and its prediction.
However, this purely autoregressive formulation is highly impractical, as it implies long encoding/decoding times.
Therefore, SReC uses a multi-resolution prediction strategy.
A low-resolution version $y^{(1)}$ of the original image $x^{(0)}$ is built through 2$\times$2 average pooling, that is
\begin{equation}
    y^{(1)}_{i,j} = \frac{x^{(0)}_{2i,2j}+x^{(0)}_{2i+1,2j}+x^{(0)}_{2i,2j+1}+x^{(0)}_{2i+1,2j+1}}{4}
\end{equation}
Then, each four-pixel group of the high-resolution image is predicted based only on the low-resolution image, independent of other groups at the same resolution level,
allowing for parallel processing and high-speed encoding.
Since the fourth pixel of a group is known, given the other three and the low resolution image, the conditional joint distribution of the group reads 
\begin{equation}
\begin{split}
    P( x^{(0)}_{2i,2j}, x^{(0)}_{2i+1,2j}, x^{(0)}_{2i,2j+1} | Y^{(1)}_{i,j}) = 
                & P( x^{(0)}_{2i,2j} | Y^{(1)}_{i,j}) \cdot
                  P( x^{(0)}_{2i+1,2j} | x^{(0)}_{2i,2j}, Y^{(1)}_{i,j}) \\
                & \cdot P( x^{(0)}_{2i,2j+1} | x^{(0)}_{2i,2j}, x^{(0)}_{2i+1,2j}, Y^{(1)}_{i,j}) 
\end{split}
\end{equation}
where $Y^{(1)}_{i,j}$ is the relevant context in the lower resolution image, that is a receptive field centered on $y^{(1)}_{i,j}$. Each term in this factorization is estimated by a dedicated convolutional neural network (CNN).
In particular, a parametric distribution is assumed, given by the mixture of $K$ discrete logistic distributions,

\begin{figure}[t!]
    \centering
    \includegraphics[page=3, width=1.00\linewidth, trim=40 120 40 0,clip]{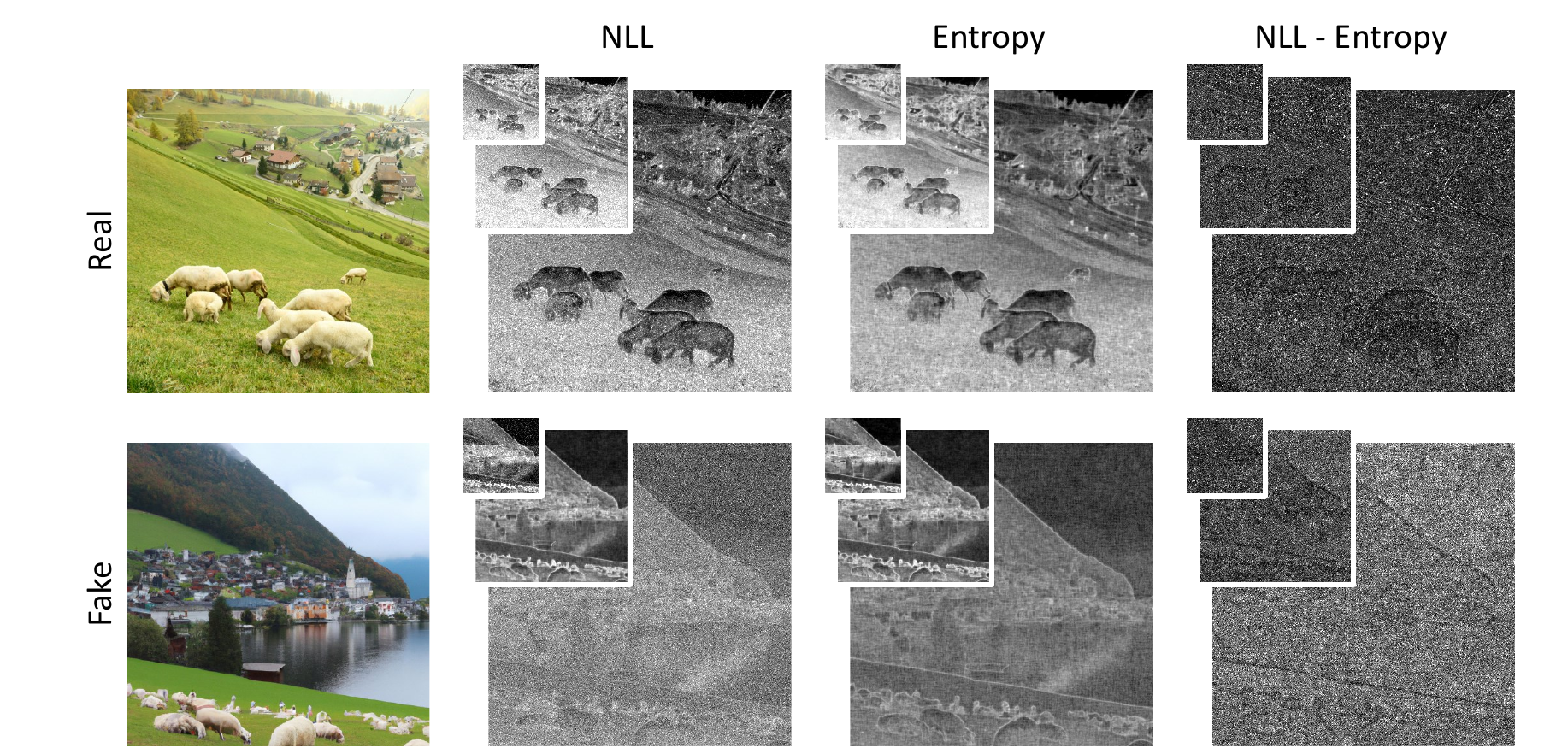}
\caption{
NLL and Entropy.
We compute the spatial distribution of NLL and Entropy at three resolutions.
For real images (top) the paired maps are very similar at all scales: when the uncertainty on a pixel (entropy) grows, also the coding cost (NLL) does. Therefore, the NLL-Entropy difference maps are all very dark.
For synthetic images (bottom) NLL and Entropy maps are not always similar, because the model is not correct, and hence the difference maps are brighter, especially the high-resolution map.
}
    \label{fig:examples}
  \vspace{-0.5cm}
\end{figure}

\begin{equation}
    P(x|X) = \sum_{k=1}^K w_k {\rm logistic}(x|\mu_k,s_k)
\end{equation}
where logistic$(x|\mu,s)=\sigma(\frac{x-\mu+0.5}{s}) - \sigma(\frac{x+\mu+0.5}{s})$ is the difference of two sigmoid functions, with position parameter $\mu$ and scale parameter $s$, and $K=10$ is always assumed.
The CNN takes the context $X$ of the pixel of interest as input and outputs the weights of the mixture together with the position and scale parameters of all logistics.
In turn, these parameters allow one to compute the desired distribution.
This whole process is replicated on two more lower-resolution scales, for a total of four levels,
the lowest resolution, an 8$\times$8 subsampled ``prompt'' image, coded in clear,
and three higher resolution images, each one predicted from its lower resolution version.
All networks are trained to minimize the cross entropy between
the predicted model probability $P_\theta(x)$ and the empirical data distribution $P(x)$ given by the training image dataset.
We mention in passing that this loss is closely related to the log-perplexity considered for text synthesis.

To summarize,
SReC provides us with a lightweight tool for computing conditional distributions of all image pixels at three different levels of resolution,
and therefore to compute all kinds of statistics that can expose the mismatch between a test image and the learned model.
Considering that SReC achieves state-of-the-art performance in lossless image compression,
one can also argue that the learned model of real images is very accurate.
Given this tool, we can now design a zero-shot detector of synthetic images.

\subsection{Features and Decision Statistics}
\label{sec:statistics}

Let $x \in \{0,\ldots,255\}^{N \times M \times 3}$ be the image under test.
In our multi-resolution framework, this will be the highest-resolution version, $x^{(0)}=x$. 
Through 2$\times$2 average pooling, we generate a lower resolution version $y^{(1)}={\rm avpool}(x^{(0)})$,
and then, through rounding, its integer-valued version $x^{(1)}={\rm round}(y^{(1)})$.
The process is repeated, and eventually we have four integer versions of the image $\{x^{(0)}, x^{(1)}, x^{(2)}, x^{(3)}\}$,
together with three non-integer versions $\{y^{(1)}, y^{(2)}, y^{(3)}\}$.
In the context of lossless coding, the lowest resolution version, $x^{(3)}$, must be sent in clear
together with the rounding bits at levels 3, 2, and 1, 
but we mention this only for completeness and for a more compelling interpretation of results.
The CNNs trained on real images provide the predicted probability distribution for all 
pixels\footnote{More precisely, all color components of all pixels, but to simplify notations, in the following we will neglect color and treat the image as if grayscale.}
of levels 0, 1, and 2
\begin{equation}
    P(x^{(l)}_{i,j}=k | X^{(l)}_{i,j})
\end{equation}
where $k \in \{0,\ldots,255\}$ and $X^{(l)}_{i,j}$ is the context for pixel $x^{(l)}_{i,j}$,
including a portion of the lower-resolution image $y^{(l+1)}$ 
and possibly some same-resolution neighbors of the current pixel.
Given the above distribution, we compute the negative log likelihood and the entropy at each pixel
\begin{eqnarray}
    \NLL^{(l)}_{i,j} & = & - \log P(x^{(l)}_{i,j} | X^{(l)}_{i,j}) \nonumber \\
    H^{(l)}_{i,j}    & = & - \sum_k P(k | X^{(l)}_{i,j}) \log P(k | X^{(l)}_{i,j}) 
\end{eqnarray}
These quantities are shown in Fig.\ref{fig:examples} for two sample images, real and synthetic.
Then, through spatial averaging, we obtain the corresponding quantities for the images at all resolution levels
$\NLL^{(l)} = \langle \NLL^{(l)}_{i,j} \rangle$ and $H^{(l)} = \langle H^{(l)}_{i,j} \rangle$, for $l=0,1,2$.
These are the features associated by the system to input image and our decision statistics will be suitable combinations of them.

\begin{figure}[t!]
    \centering
    \includegraphics[page=2, width=1.00\linewidth, trim=0 240 0 0,clip]{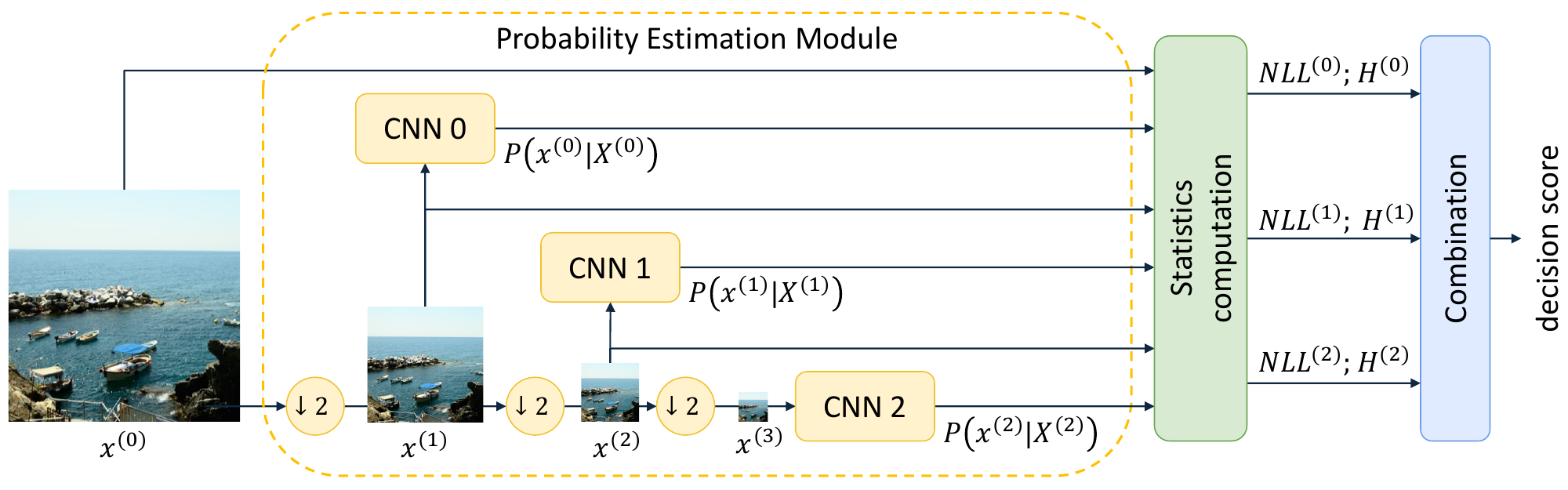}
    \caption{
Extracting decision statistics.
The full resolution image $x^{(0)}$ is downsampled three times.
The lowest-resolution version, $x^{(3)}$, feeds the level-2 CNN, which outputs the probability distributions of level-2 pixels.
These distributions, together with the actual level-2 pixels, are used to compute the level-2 coding cost ${\rm NLL}^{(2)}$ and its expected value $H^{(2)}$.
All these steps are then repeated for levels 1 and 0.
Eventually, NLLs and entropies are combined to compute the decision statistics. 
} 
\label{fig:schema}
\vspace{-0.5cm}
\end{figure}

Before going on, it is convenient to give a physical interpretation of these quantities.
Each $\NLL$ can be interpreted as the actual coding cost for the corresponding image.
While each entropy can be interpreted as the expected value of the coding cost given the context,
when the image is coherent with the predicted distribution.
In the presence of a mismatch, $\NLL-H>0$, on the average, with a gap that increases with increasing distribution mismatch.
Our fundamental assumption is that the trained CNNs provide a good model of real images, and synthetic images tend not to follow the same model.
Therefore, we expect that synthetic images are characterized by higher coding cost, hence higher NLL, under this distribution.
This observation would lead us to use the NLLs as decision statistics.
However, the coding cost does not depend only on the distribution mismatch but also (predominantly) on the intrinsic information content of the image, measured by the entropy. 
A complex image, say a photo of a crowd, is more difficult to encode/describe than a smooth image, say a blue sky, no matter what model we use.
Therefore, to get rid of this bias, we consider the coding cost gap, defined as the difference $D^{(l)}=\NLL^{(l)}-H^{(l)}$, as decision statistic.
Hence, for each image, we have three basic decision statistics, one for each resolution level.
It is worth observing that some forms of normalization are adopted for machine generated text detection as well \cite{mitchell2023detectGLPT,su2023detectllm,hans2024spotting}. A block diagram of our method is shown in Fig.\ref{fig:schema}.

A sample graph of the coding cost gap is shown in Fig.\ref{fig:teaser}, on the right.
For real images and three families of synthetic images we report the average gap (solid line) plus/minus its standard deviation (colored band) for the various resolutions levels. 
Two important observations can be made.
First of all, the level-0 coding cost gap, concerning the full resolution image, seems to be much more discriminant than the others. Moreover, the gap grows much faster for synthetic images than for real images when going from level 1 to level 0. 
Therefore, as decision statistics we will consider both $D^{(0)}$ (the level-0 coding cost gap)  and  $\Delta^{01}=D^{(0)} - D^{(1)}$ (its slope). 
In addition, in preliminary experiments we observed that synthetic images are sometimes characterized by a coding cost much {\em lower} rather than much higher than expected, that is the NLL is much lower than the entropy. This is also an anomaly, which signals the likely synthetic nature of the image.
Therefore, besides the above statistics we also consider their absolute values $\left| D^{(0)} \right|$ and $\left|  \Delta^{(01)} \right|$.
These observations are supported by the sample graphical analysis shown in Fig.\ref{fig:scatters} in the ablation study.

\section{Results}
\label{sec:results}

\subsection{Datasets and Metrics}
\label{sec:datasets}

\begin{figure}[t!]
    \centering
    \includegraphics[page=4,width=0.80\linewidth]{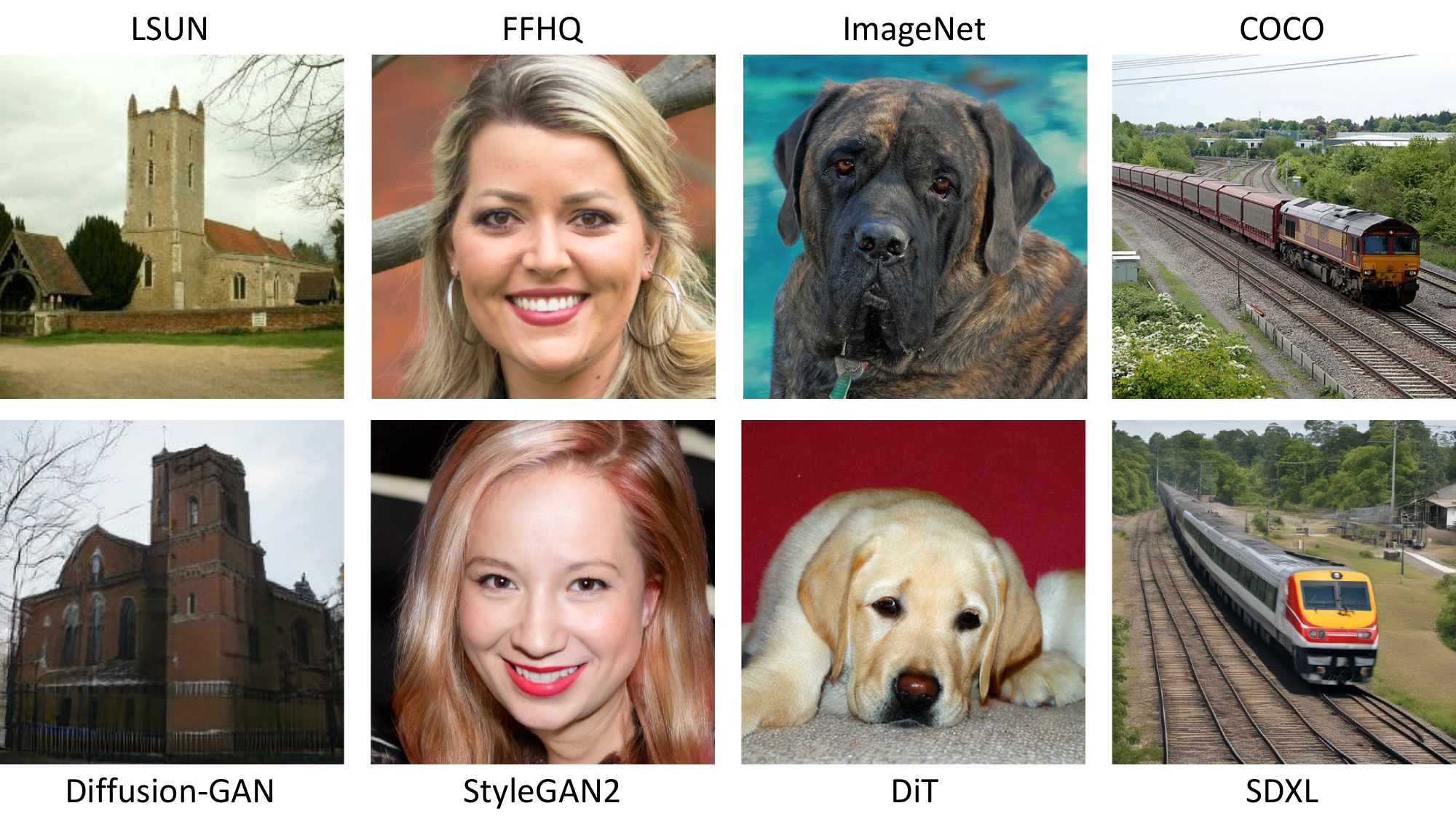}
    \caption{Examples of real and AI-generated images of different categories used in our experiments. Top: real images from LSUN, FFHQ, ImageNET and COCO. Bottom: generated images from DiffusionGAN, StyleGAN2, DiT and SDXL.}
    \label{fig:images}
\end{figure}

We benchmarked our model on a large variety of synthetic generators both GANs and DMs:
 GauGAN \cite{park2019semantic},
 BigGAN \cite{brock2018large},
 StarGAN \cite{choi2018stargan},
 StyleGAN2 \cite{karras2020stylegan2},
 Diffusion-GAN   \cite{wang2023diffusion},
 GigaGAN \cite{kang2023scaling},
 GALIP \cite{tao2023galip},
 DDPM \cite{ho2020denoising},
 ADM \cite{dhariwal2021diffusion},
 GLIDE \cite{nichol2021glide},
 Stable Diffusion \cite{stablediffusion1,stablediffusion2},
 DiT \cite{peebles2022scalable},
 DeepFloyd-IF \cite{deepfloydif},
 Stable Diffusion XL \cite{podell2023sdxl},
 DALL·E \cite{Dayma_DALLE_Mini_2021},
 DALL·E 2 \cite{ramesh2022hierarchical},
 DALL·E 3 \cite{dalle3},
 Midjourney V5 \cite{midjourney},
 and Adobe Firefly \cite{firefly}.
We collected images from publicly available datasets \cite{wang2020cnn, corvi2023detection, ojha2023towards, bammey2023synthbuster} and generated additional images as needed when they were not publicly available.
We ensured that all datasets included pristine and synthetic images with similar semantic content, both compressed and uncompressed, to avoid any kind of bias (see Fig.\ref{fig:images}). 
For some synthetic generators we have multiple datasets, built on the basis of different real image datasets
LSUN \cite{yu2015lsun}, FFHQ \cite{karras2019stylegan}, ImageNet \cite{deng2009imagenet}, COCO \cite{lin2014microsoft}, LAION \cite{schuhmann2021laion} and RAISE \cite{nguyen2015raise}.
This is a fortunate circumstance: we kept them carefully separate as this allows us to analyze how the performance of a detector depends on the class of real images used in the synthesis phase.
Overall we used a total of 29k synthetic images and 6k real images.
More details on the generated and actual images are provided in the supplementary material.

Following other papers \cite{liu2022detecting,ojha2023towards,cozzolino2023raising} we measure performance using the area under the ROC curve (AUC) and the balanced accuracy. We also show the influence of the threshold selection on the performance.

\subsection{Ablation Study}
\label{sec:ablation}

\subsubsection{Features analysis.} 
\begin{figure}[t!]
    \centering
    \includegraphics[page=1, height=0.3\linewidth, trim=0 0 0 0,clip]{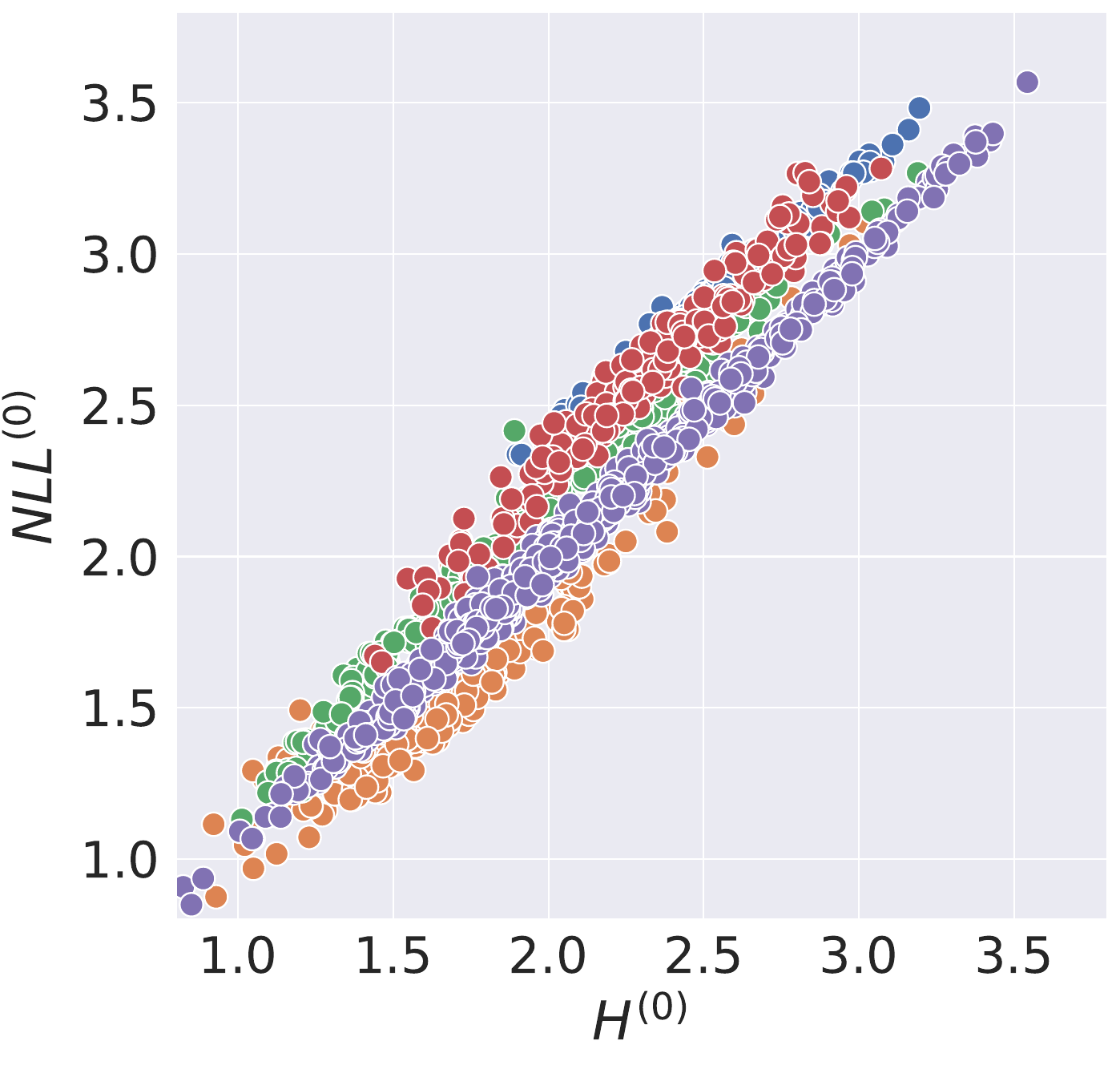}
    \includegraphics[page=2, height=0.3\linewidth, trim=0 0 0 0,clip]{figures/scatter3.pdf}
    \includegraphics[page=3, height=0.3\linewidth, trim=0 0 0 0,clip]{figures/scatter3.pdf} \\
    \includegraphics[page=4, width=0.7\linewidth, trim=0 0 0 10,clip]{figures/scatter3.pdf}
    \caption{
Decision statistics.
NLL and entropy by themselves are not discriminant (left).
Their difference (center) is much more useful for detection, 
but only at high resolution, $D^{(0)}$, while $D^{(1)}$ is less discriminant and $D^{(2)}$ basically useless.
Right box shows histograms of $D^{(0)}$ for real and synthetic images.
Note that for GLIDE, $D^{(0)}$ is negative, on the average.
Good discrimination is still possible based on the absolute value.
}
    \label{fig:scatters}
\end{figure}

First, we want to provide a better insight into the role and importance of the features described in Section 3.4:
$D^{(0)}$ (the 0-level coding cost gap),
its slope $\Delta^{01} = D^{(0)} - D^{(1)}$
and their absolute values.
To this end, we consider the set of real and synthetic (DALL·E 2, GLIDE, Midjourney, SDXL) images of the Synthbuster dataset \cite{bammey2023synthbuster}.
We note, in passing, that this dataset includes only uncompressed images, which dispels any possible doubt that our method exploits some JPEG compression bias between real and fake images \cite{grommelt2024fake}.
Some selected scatter plots and graphs are shown in Fig.\ref{fig:scatters}.
The rightmost box shows that encoding cost (NLL) and entropy ($H$) alone are not very discriminating,
even if computed at the more informative level 0 (high resolution).
In contrast, their difference, the 0-level coding cost gap $D^{(0)}$,
seems to separate the different classes quite well (central box), in particular the real class (violet) from the others.
Note that the level-1 gap (not shown) is not equally discriminating,
and the level-2 gap, plotted on the $y$ axis, turns out to be essentially useless.
In the third box we plot the empirical distributions of $D^{(0)}$ for the various classes.
This representation makes the good separability of the classes further clear but also highlights an unexpected phenomenon: 
GLIDE images group mostly to the left of the real class, that is, they have a lower-than-expected coding cost.
Although not in line with our initial hypotheses, this fact nevertheless represents an anomaly, 
which can be detected by thresholding the absolute value of the statistic rather than the statistic itself.

\begin{figure}[t!]
    \centering
    \includegraphics[page=3,width=0.9\linewidth]{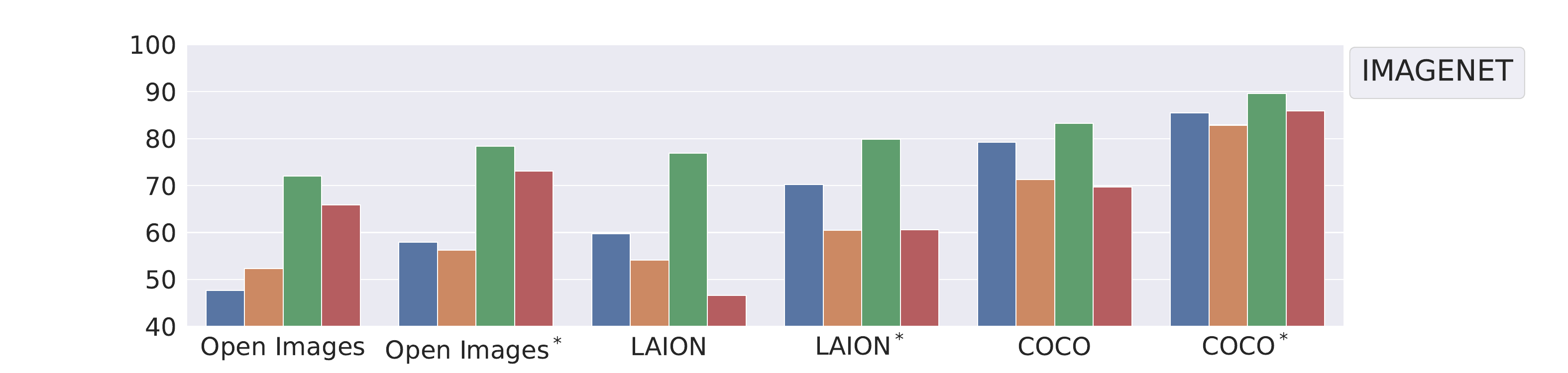} 
    \includegraphics[page=6,width=0.9\linewidth]{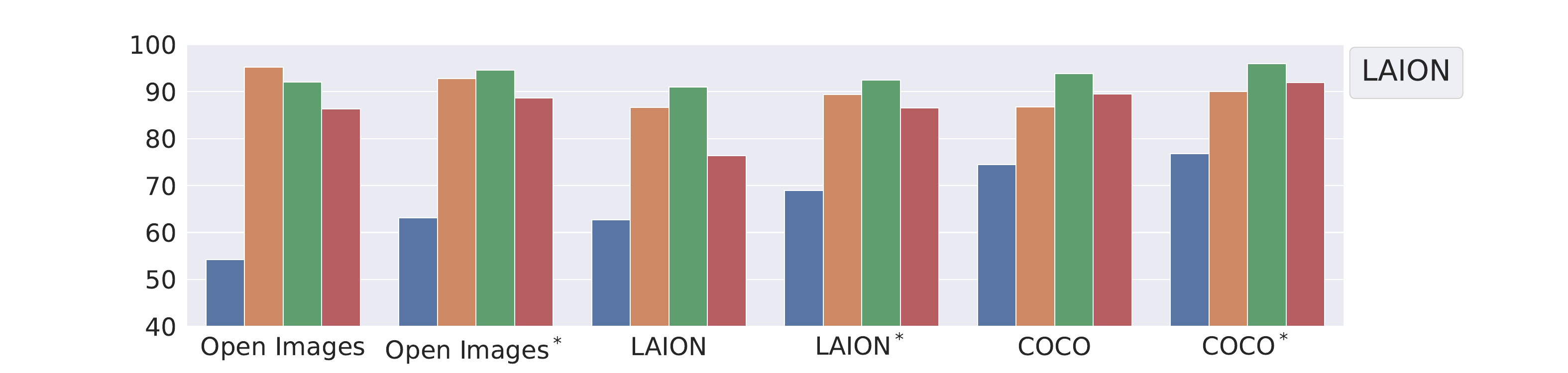} \\
    \includegraphics[page=7,width=0.5\linewidth]{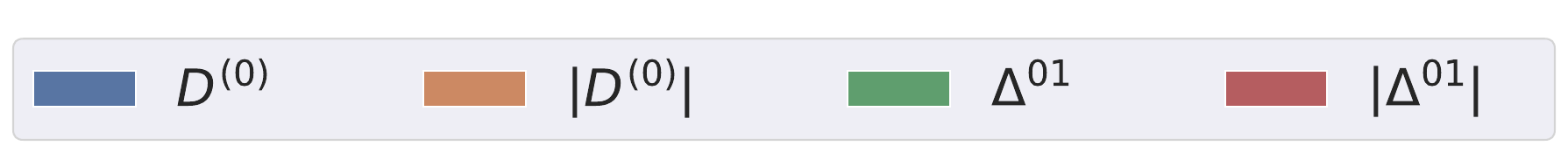} 
    \caption{AUC of proposed method as a function of decision statistic (see Section \ref{sec:statistics}) and dataset of real images used to train the lossless encoder: Open Images, LAION, COCO, and their augmented versions ($^*$).
    Synthetic test images are selected to match the corresponding real test images: ImageNet (top), and LAION (bottom).}
    \label{fig:ablation}
\end{figure}

\vspace{0mm} 
\subsubsection{Influence of the real class.}
To better understand the role of the real dataset used to train the lossless encoder, we perform an experiment in which we vary it.
Along with the original encoder pre-trained on the Open Images dataset \cite{krasin2017openimages} (about 338k high-resolution images), we consider two other versions, trained from scratch on the LAION dataset \cite{schuhmann2021laion} ($\simeq$117k), and the COCO dataset \cite{lin2014microsoft} ($\simeq$106k), respectively, using the same hyperparameters as \cite{cao2020lossless}.
Additionally, we consider versions (marked with *) trained on the same datasets, augmented with JPEG compressed images with quality between 80 and 100.
We compute the performance in terms of AUC on two different datasets of synthetic and real images, where this latter class comes from ImageNet \cite{deng2009imagenet} (Fig.\ref{fig:ablation}, top) or LAION \cite{schuhmann2021laion} (Fig.\ref{fig:ablation}, bottom).
We can observe that the best and more uniform results across the four decision statistics are obtained using COCO$^*$, while training on Open Images guarantees 
good performance  if the real class is LAION, but bad performance if it is ImageNet. Additional results are included in the supplementary material.

\newcommand{\mc}[1]{\multicolumn{6}{l}{\rule{0mm}{5mm} {\bf #1}}\\ \hline}
\setlength{\tabcolsep}{3pt}
\begin{table}[t!]
\caption{Reference methods. 
For each one we indicate the key idea, the datasets of real and synthetic images used for training with their sizes, whether or not augmentation is used, the test strategy.}
\centering
\resizebox{\textwidth}{!}
{\scriptsize
\begin{tabular}{rccccc}
\toprule
\cmidrule(lr){3-4} \cmidrule(lr){5-6}
\ru Acronym [ref]                         & Idea/Approach     & Training Real/Fake & Size(K) & Augment.   & Test Strategy   \\ \midrule
\ru Wang2020   \cite{wang2020cnn}         & High diversity    & LSUN/ProGAN        & 360/360 & \checkmark & global pooling  \\
\ru PatchFor.  \cite{chai2020what}        & Patch-based       & CelebA,FF/various  &  84/272 &            & resizing        \\
\ru Liu2022    \cite{liu2022detecting}    & Noise-based       & LSUN/ProGAN        & 360/360 & \checkmark & global pooling  \\
\ru Corvi2023  \cite{corvi2023detection}  & No-downsampling   & COCO,LSUN/Latent   & 180/180 & \checkmark & global pooling  \\
\ru LGrad      \cite{tan2023learning}     & Gradient-based    & LSUN/ProGAN        &  72/72  & \checkmark & resizing        \\
\ru DIRE       \cite{wang2023dire}        & Inversion         & LSUN-Bed/ADM       &  40/40  &            & resizing        \\
\ru DE-FAKE    \cite{sha2022fake}         & Prompt-based      & LSUN/Stable Diff.  &  20/20  &            & resizing        \\
\ru Ojha2023   \cite{ojha2023towards}     & CLIP              & LSUN/ProGAN        & 360/360 & \checkmark & cropping        \\
\ru NPR     \cite{tan2023rethinking}      & Residual          & LSUN/ProGAN        &  72/72  &            & resizing        \\
\ru AEROBLADE  \cite{ricker2024aeroblade} & AE rec. error     &  - / -             &  - / -  &            & global distance \\
\bottomrule
\end{tabular}
}
\label{tab:summary}
\end{table}

\subsection{SoTA Comparison}
\label{sec:comparison}

In our analysis we include only methods with code and/or pre-trained models publicly available on-line. 
Eventually, we included 7 CNN-based methods \cite{wang2020cnn, chai2020what, liu2022detecting, corvi2023detection, tan2023learning, wang2023dire, tan2023rethinking}, 2 CLIP-based methods \cite{sha2022fake,ojha2023towards} and a training-free method \cite{ricker2024aeroblade}. 
A brief summary of these techniques is provided in Tab.\ref{tab:summary}, while a more detailed description is given in the supplementary material.
For a fair comparison we avoid testing on ProGAN \cite{karras2018progressive} and Latent Diffusion \cite{rombach2022high}, 
because a good number of these supervised methods were trained on datasets that include images from these generators.
Even so, we have a total of 30 datasets for testing.
Results are reported in Tab.\ref{tab:comparative_dm} in terms of AUC, with the best figure for each dataset highlighted in bold.
Note that each row is characterized by the name of the generator (e.g., GauGAN) 
and by a single letter that recalls the set of real images used to train it: S for LSUN, F for FFHQ, I for ImageNet, C for COCO, L for LAION, R for RAISE.
This detail allows us to study how the performance depends on the real dataset (but with synthetic images from the same generator and with semantic content aligned with real images).

\setlength{\tabcolsep}{4pt}
\begin{table}[!t]
    \caption{AUC for reference and proposed methods. Best score in bold with a 0.5\% margin. 
    S = LSUN, F = FFHQ, I = ImageNet, C = COCO, L = LAION, R = RAISE.}
    \label{tab:comparative_dm}
    \centering
    \resizebox{0.90\textwidth}{!}{\begin{tabular}{lc !{\color{tabvline}\vrule} cccccccccc !{\color{tabvline}\vrule} ccccc}
    \toprule
 & \rotatebox{90}{Real data} & \rotatebox{90}{Wang2020} & \rotatebox{90}{PatchFor.} & \rotatebox{90}{Liu2022} & \rotatebox{90}{Corvi2023} & \rotatebox{90}{LGrad} & \rotatebox{90}{DIRE} & \rotatebox{90}{DEFAKE} & \rotatebox{90}{Ojha2023} & \rotatebox{90}{NPR} & \rotatebox{90}{AEROBLADE} & \rotatebox{90}{Ours $D^{(0)}$ } & \rotatebox{90}{Ours $\left|  D^{(0)} \right|$} & \rotatebox{90}{Ours $\Delta^{01}$} & \rotatebox{90}{Ours $\left| \Delta^{01} \right|$} \\
\midrule 
\band          GauGAN & C &    98.9  &    80.8  & \b{99.7} &    83.8  &    81.6  & \b{99.9} &    43.8  & \b{100.} &    89.1  &    55.1  & \b{99.8} & \b{99.8} & \b{99.9} & \b{99.7} \\
          BigGAN & I &    92.7  &    85.5  &    94.7  &    83.4  &    77.2  & \b{99.8} &    59.0  & \b{99.6} &    86.8  &    51.9  &    92.3  &    88.6  &    95.9  &    92.6  \\
\band         StarGAN & F &    94.7  & \b{100.} & \b{99.9} &    95.9  &    73.9  &    40.4  &    45.9  & \b{99.7} &    81.5  &    84.0  & \b{100.} & \b{100.} & \b{100.} & \b{100.} \\
                 & S &    98.1  &    83.8  & \b{99.7} &    89.1  & \b{99.8} &    58.3  &    39.1  &    96.7  & \b{100.} &    30.0  &    96.6  &    96.1  &    96.7  &    96.5  \\
 \multirow{-2}{*}{StyleGAN2} & F &    94.9  &    85.1  & \b{99.9} &    58.4  &    82.7  &    55.5  &    47.6  &    91.0  &    71.3  &    60.1  &    43.1  &    87.7  &    41.1  &    88.7  \\
\band                 & I &    73.7  &    61.0  &    97.3  &    50.5  &    76.4  & \b{99.9} &    64.3  &    94.6  &    82.4  &    47.5  &    72.4  &    68.1  &    72.4  &    68.1  \\
\band \multirow{-2}{*}{GigaGAN} & C &    79.5  &    84.0  & \b{99.6} &    90.9  &    76.7  & \b{99.9} &    87.9  &    97.6  &    95.5  &    80.6  &    96.5  &    94.0  &    96.7  &    93.4  \\
        Diff.GAN & S &    89.8  &    92.6  &    99.5  &    96.6  & \b{99.5} &    49.8  &    44.8  &    97.4  & \b{100.} &    43.9  &    99.4  &    99.4  &    99.5  &    99.5  \\
\band           GALIP & C &    89.7  &    98.2  &    94.3  &    87.7  &    56.7  & \b{100.} &    75.6  &    98.6  &    90.7  &    65.0  &    98.4  &    96.3  & \b{99.7} &    99.4  \\
          DALL·E & L &    66.4  &    71.7  &    95.0  &    98.3  &    95.2  & \b{99.8} &    55.9  &    97.3  & \b{99.5} &    24.1  &    99.2  &    95.8  &    98.2  &    95.4  \\
\band            DDPM & F &    31.6  &    98.4  &    22.8  & \b{100.} &     9.8  &    23.1  &    50.5  &    77.7  &    92.4  &    81.7  &    76.6  &    25.2  &    93.8  &    79.6  \\
                 & S &    67.6  &    67.6  &    70.6  &    80.3  &    81.1  &    52.0  &    37.4  &    88.2  & \b{94.1} &    53.1  &    49.5  &    53.5  &    69.4  &    71.0  \\
 \multirow{-2}{*}{ADM} & I &    61.0  &    81.9  &    94.4  &    81.1  &    72.7  & \b{99.5} &    69.1  &    85.3  &    78.5  &    80.3  &    87.8  &    90.5  &    95.3  &    92.1  \\
\band                 & C &    64.8  &    97.4  &    96.3  &    97.2  &    81.5  & \b{99.9} &    92.4  &    88.8  &    95.4  &    98.0  &    47.8  &    88.5  &    91.1  &    91.1  \\
\band                 & R &    32.2  & \b{95.0} &    56.6  &    86.5  &    50.6  &    42.9  &    92.2  &    72.8  &    63.3  &    87.7  &    23.2  &    89.4  &    51.1  &    65.1  \\
\band \multirow{-3}{*}{GLIDE} & L &    72.6  &    74.1  &    90.8  &    86.9  &    90.3  & \b{100.} &    60.2  &    95.3  & \b{99.8} &    68.7  &    54.5  &    84.2  &    93.8  &    88.5  \\
             DiT & I &    58.6  &    83.1  &    88.0  & \b{100.} &    56.2  & \b{99.6} &    87.4  &    77.8  &    78.4  & \b{99.8} &    89.4  &    84.3  &    94.9  &    91.0  \\
\band                 & C &    68.2  &    86.1  &    95.3  & \b{100.} &    54.7  & \b{99.9} &    93.3  &    97.9  &    76.5  & \b{99.8} &    48.4  &    74.8  &    54.6  &    71.4  \\
\band \multirow{-2}{*}{Stable D. 1.4} & R &    37.9  &    61.8  &    73.4  & \b{100.} &    50.0  &    37.6  &    88.0  &    87.7  &    43.0  &    96.9  &    99.4  &    98.7  &    97.0  &    97.2  \\
                 & C &    56.5  &    78.6  &    94.2  & \b{100.} &    62.8  &    99.3  &    97.9  &    82.3  &    89.3  & \b{99.9} &    83.0  &    90.3  &    84.5  &    89.1  \\
 \multirow{-2}{*}{Stable D. 2} & R &    50.2  &    38.7  &    34.8  & \b{100.} &    41.4  &    35.5  &    80.7  &    89.5  &    44.0  &    97.4  &    98.5  &    96.8  &    95.8  &    95.9  \\
\band                 & C &    83.8  &    60.8  &    89.3  & \b{100.} &    89.3  &    99.5  &    94.0  &    80.0  &    99.3  &    87.9  & \b{99.9} & \b{99.9} & \b{99.9} & \b{99.8} \\
\band \multirow{-2}{*}{SDXL} & R &    54.3  &    68.4  &    31.1  & \b{100.} &    57.2  &    47.1  &    84.4  &    85.1  &    76.7  &    69.7  & \b{100.} & \b{100.} &    99.1  &    99.2  \\
        Deep.-IF & C &    78.0  &    62.7  &    72.2  & \b{99.9} &    68.8  &    98.9  &    96.9  &    92.9  &    91.6  &    81.9  &    91.7  &    82.3  &    88.4  &    79.4  \\
\band                 & C &    88.5  &    52.4  &    98.9  &    88.2  &    78.6  & \b{99.9} &    80.6  &    97.1  &    90.0  &    59.3  & \b{100.} & \b{100.} & \b{100.} & \b{99.9} \\
\band \multirow{-2}{*}{DALL·E 2} & R &    64.8  &    41.9  &    70.4  &    69.4  &    58.6  &    44.7  &    70.9  &    95.2  &    39.5  &    32.8  & \b{100.} & \b{100.} & \b{100.} & \b{100.} \\
                 & C &    65.0  &    47.3  &    99.5  & \b{100.} &    88.4  & \b{99.9} &    96.2  &    86.4  &    97.7  & \b{99.7} & \b{99.7} &    99.5  &    98.3  &    98.2  \\
 \multirow{-2}{*}{DALL·E 3} & R &    10.9  &    52.7  &     0.2  &    60.8  &    37.9  &    47.6  & \b{92.4} &    36.4  &    48.7  &    48.3  &    79.1  &    66.7  &    78.0  &    78.1  \\
\band      Midjourney & R &    40.2  &    57.8  &    40.7  & \b{100.} &    56.3  &    51.0  &    78.1  &    66.2  &    77.0  &    99.0  & \b{99.7} &    99.3  &    98.5  &    98.5  \\
   Adobe Firefly & R &    84.8  &    49.4  &    11.8  & \b{98.0} &    40.6  &    57.4  &    81.4  & \b{97.5} &    32.1  &    52.8  &    73.6  &    41.2  &    80.8  &    80.4  \\
\midrule 
\band             AVG &   &    68.3  &    73.3  &    77.0  &    89.4  &    68.2  &    74.6  &    72.9  &    88.4  &    80.1  &    71.2  &    83.3  &    86.4  &    88.8  &    \b{90.0}  \\
\bottomrule
    \end{tabular}}
\end{table}
\setlength{\tabcolsep}{1.4pt}

First of all, we observe that for most reference methods the average AUC does not exceed 80\%.
Notable exceptions are the CLIP-based Ojha2023 (88.4\%) and the CNN-based Corvi2023 (89.4\%).
Interestingly, some methods show very different performance when the real class changes. This may be due to JPEG bias as already suggested in \cite{grommelt2024fake, ricker2024aeroblade}. A deeper analysis on this point is presented in the supplementary material.
The proposed zero-shot approach goes above 80\% with all decision statistics, reaching the top value of 90.0\% when $|\Delta^{01}|$ is used.
Obviously, this is a very good result, 
but what makes it especially valuable is the absence of any dependence on the generators' models.
This point is further stressed by the fact that the AUC remains extremely stable across all test sets, with a minimum of 65.1\% on GLIDE-R.
On the contrary, the best competitor, Corvi2023, has a long score of top results but also some very poor ones. 
suggesting a certain instability, likely due to the presence/absence of specific artifacts in the test images,
and eventually the risk of not adapting to models of new conception.
We want also to draw the reader's attention on the already mentioned case of GLIDE and on the fact that 
the proposed method exhibits wildly different results with different decision statistics.
In particular, with $|D^{(0)}|$ the AUC is 89.4\% as opposed to the already mentioned 65.1\% with $|\Delta^{01}|$.
This suggests there may be better ways to exploit the basic NLL$^{(l)}$ and $H^{(l)}$, possibly jointly at all levels, 
to synthesize a better and more stable decision statistics.

\begin{figure}[t!]
    \centering
    \includegraphics[page=4,trim = 25 0 0 0, clip,width=0.24\linewidth]{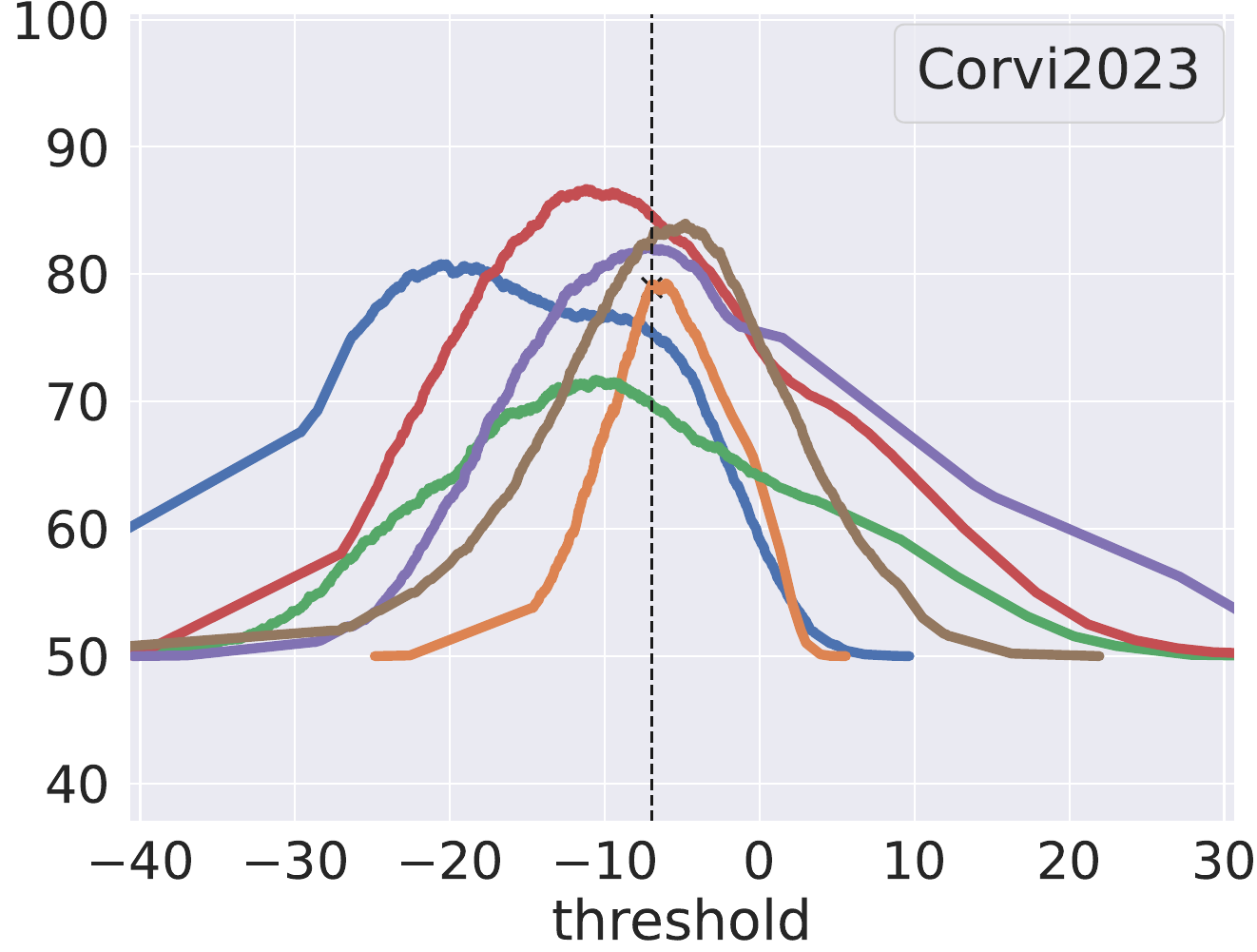} 
    \includegraphics[page=8,trim = 25 0 0 0, clip,width=0.24\linewidth]{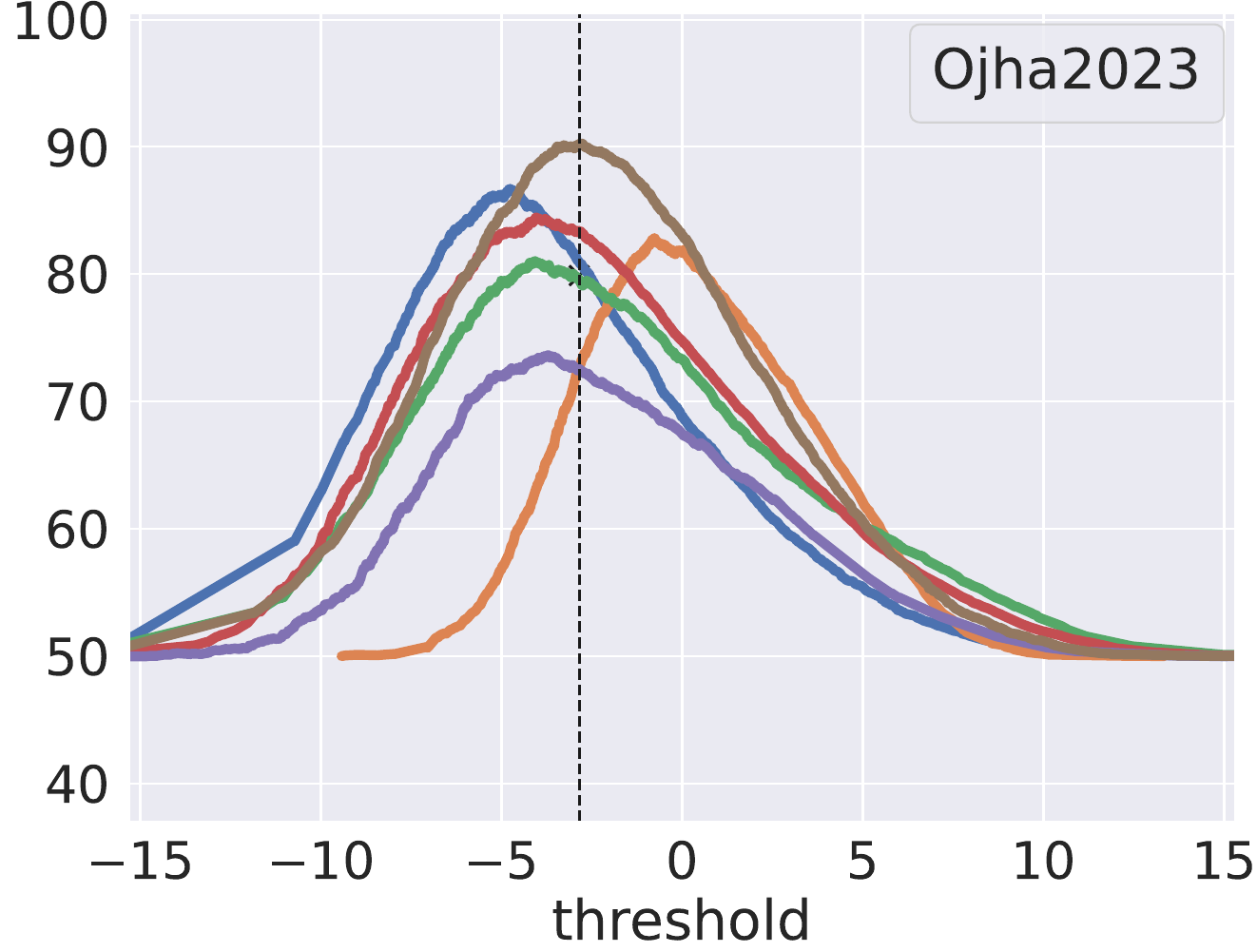}
    \includegraphics[page=9,trim = 25 0 0 0, clip,width=0.24\linewidth]{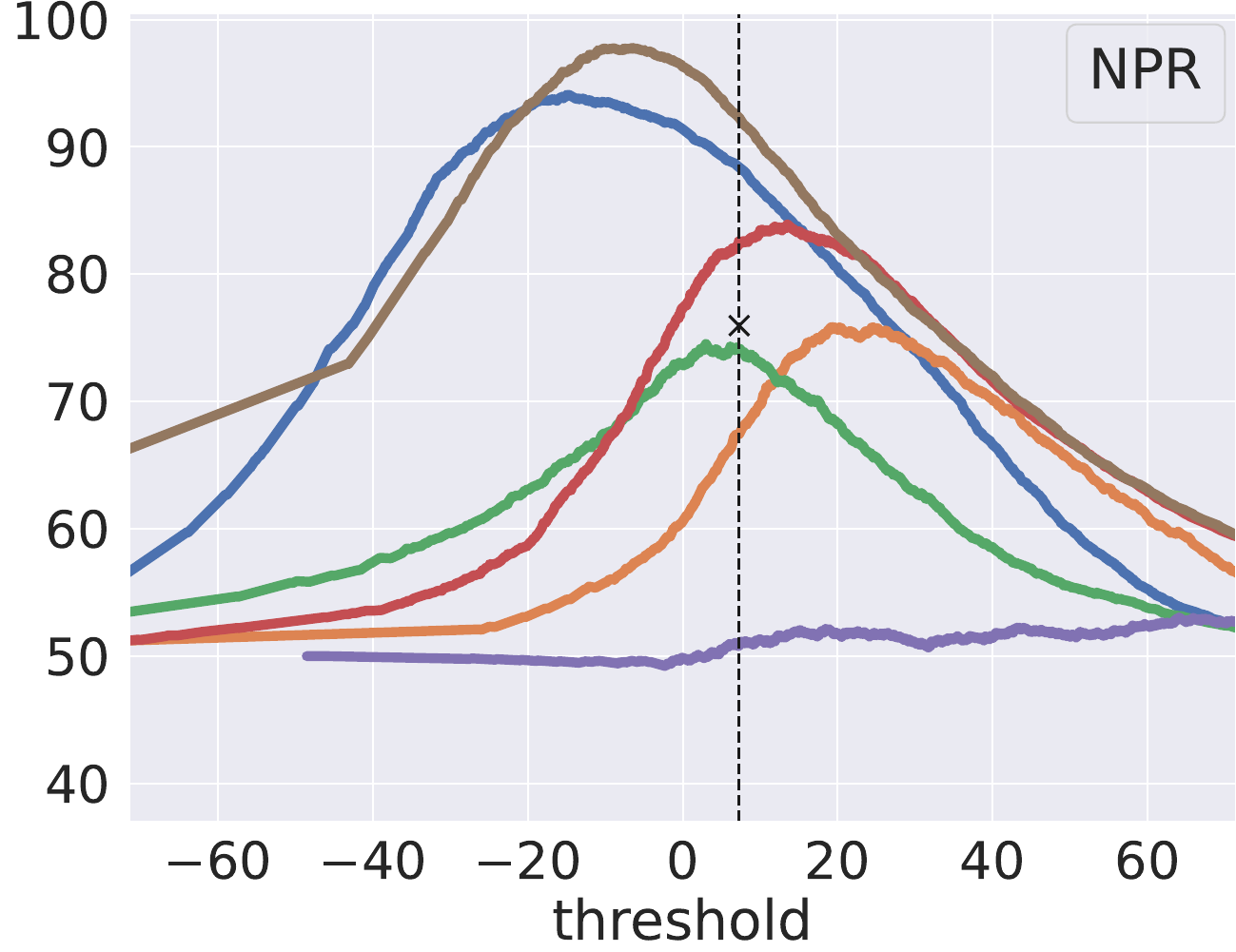} 
    \includegraphics[page=11,trim = 25 0 0 0, clip,width=0.24\linewidth]{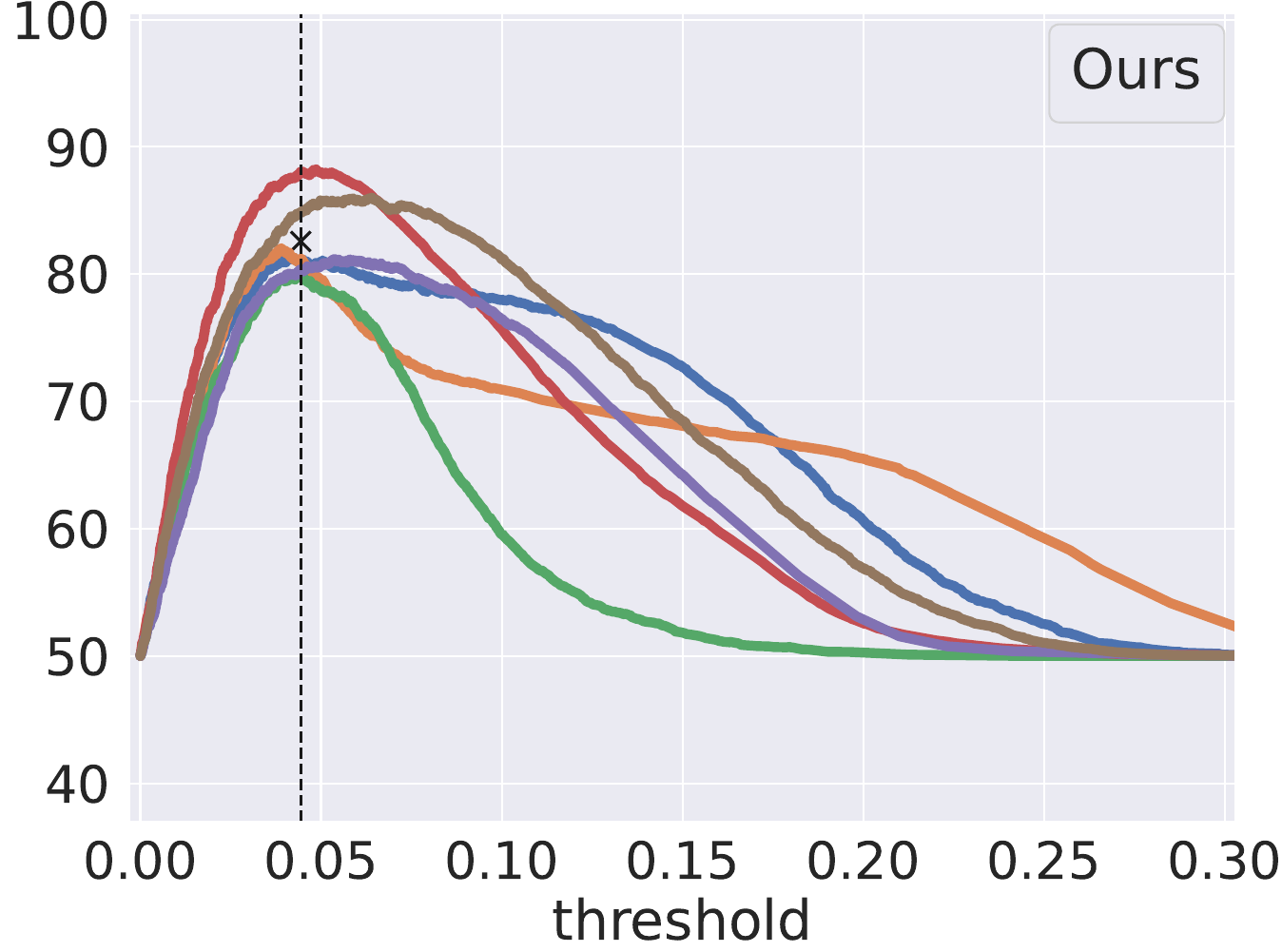}  \\
    \includegraphics[page=12,width=0.7\linewidth]{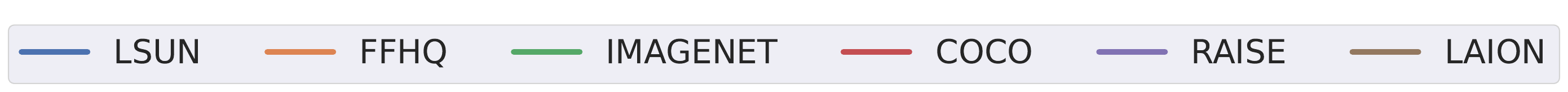}
\caption{
Balanced accuracy as a function of the detection threshold.
For each dataset of real images, we average accuracy over all associated synthetic generators.
The dotted vertical line indicates the global optimal threshold and the $\times$ symbol the corresponding accuracy. 
Note that only for the proposed method all peaks are very close, indicating the presence of a single threshold. Charts for other methods are reported in the Suppl.
}
    \label{fig:graph_ths}
\end{figure}

Finally, in Fig.\ref{fig:graph_ths}, we report the accuracy as a function of the decision threshold for the best methods.
A separate curve is shown for each real image dataset by averaging over the associated synthetic generators.
Unlike AUC, the accuracy critically depends on the selection of a good threshold and some calibration data may be needed for this purpose.
Note that only for the proposed method there is a single good threshold that ensures near-optimal accuracy for all datasets.

\subsection{Limitations}
Our work was developed to detect whether an image has been fully generated and not to detect local manipulations. 
However, it could be easily extended to accomplish this task since we already compute a map of local pixel-wise statistics.
Furthermore, our approach relies on a model of the real class learned by the encoder. If real images do not satisfy this model, the approach may not perform correctly. For example, if images are highly compressed or resized (as is the case on the web), statistical analysis may not be reliable.

\section{Conclusion}
\label{sec:discussion}

We introduced a novel zero-shot forensic detector to distinguish AI-generated images from real ones.
Unlike most current methods, our approach does not require fake images during training, which ensures generalization to yet unknown generative models.
The idea is to exploit an implicit model of real images and classify off-model images as synthetic.
To this end, we leverage an appropriate lossless encoder, trained only on real images, that can predict the probability distribution of each pixel given its context.
Synthetic images are expected to not respect this distribution, thus revealing their artificial nature.
Our experiments show that the proposed detector is consistently competitive with detectors trained in supervised modality, and outperforms them in terms of generalization ability.
We believe that our approach is an important stepping stone towards effective forensic tools that can operate without relying on domain- or method-specific training data. 
Future work will focus on making the method robust to the most common forms of image impairment, so as to make it suitable for in the wild application.

\subsubsection{Acknowledgments.}
We gratefully acknowledge the support of this research by a TUM-IAS Hans Fischer Senior Fellowship, the ERC Starting Grant Scan2CAD (804724), and a Google Gift.
This material is also based on research sponsored by the Defense Advanced Research Projects Agency (DARPA) and the Air Force Research Laboratory (AFRL) under agreement number FA8750-20-2-1004.
The U.S. Government is authorized
to reproduce and distribute reprints for Governmental purposes notwithstanding any copyright notation thereon. The
views and conclusions contained herein are those of the
authors and should not be interpreted as necessarily representing the official policies or endorsements, either expressed or implied, of DARPA or the U.S. Government.
In addition, this work has received funding by the European Union under the Horizon Europe vera.ai project, Grant Agreement number 101070093.

\begin{appendix}
\section*{Supplemental Material}
In this appendix,
we give more details about the comparison methods and the evaluation datasets present in the main paper. Moreover, we report additional ablation results and experimental analysis.

\section{Reference methods} 
In our comparative analysis, in order to ensure a fair comparison, we include only reference techniques with code and/or pre-trained models publicly available on-line.
Eventually, we considered the following methods, listed approximately from less recent to most recent:

\begin{enumerate}

\item {\bf Wang2020}~\cite{wang2020cnn}
it is based on a plain ResNet50 backbone and represents a reference in the research community.
Its main peculiarity is the training phase,
based on a large dataset of 360k real (LSUN) and 360k synthetic (ProGAN) images, the latter taken from 20 different classes.
Augmentation based on compression and blurring is also used.
The proposed dataset has been widely adopted in subsequent papers to train new models.

\vspace{2mm}
\item {\bf PatchFor}~\cite{chai2020what}
is a fully convolutional classifier on an XceptionNet backbone, with a limited receptive field that allow to focus on image local patches rather than on the global structure. The patch-based predictions are also used to visualize patterns that indicate the regions where real and fake image can be easily detected. Finally, images are properly pre-processed to avoid learning image formatting artifacts.
    
\vspace{2mm}
\item {\bf Liu2022}~\cite{liu2022detecting}
tries to distinguish real images from synthetic images based on statistical differences observed in their learned noise patterns. 
To improve performance, spatial information is then combined with information gathered in the frequency domain.

\vspace{2mm}
\item {\bf Corvi2023}~\cite{corvi2023detection}
it relies on a modified ResNet50 architecture where sub-sampling steps are mostly avoided to preserve forensic traces.
Also resizing is not carried out both in the training and testing phases to preserve  subtle forensic traces.
It is trained on a dataset of latent diffusion images, with strong augmentation to gain higher robustness and generalization ability.

\vspace{2mm}
\item {\bf LGrad}~\cite{tan2023learning} 
relies exclusively on low-level traces and thus removes the image content altogether, by extracting only the image gradients through a pre-trained CNN model.
In the noise residual domain, real and synthetic images are distinguished based on their different inter-pixel dependencies.

\vspace{2mm}
\item{\bf DIRE}~\cite{wang2023dire} 
is an inversion-based method, relying on the assumption that synthetic images can be well approximated by a generator while real images cannot.
To perform detection, a ResNet50 backbone is fed with the difference between the image and its version obtained by the inversion process.

\vspace{2mm}
\item{\bf DE-FAKE}~\cite{sha2022fake}
leverages an image captioning model (BLIP) to generate the textual prompt of the image.
The high-level embeddings of image and text, extracted using a large pre-trained model (CLIP), are then concatenated and fed to a multilayer perceptron for the binary classification.

\vspace{2mm}
\item{\bf Ojha2023}~\cite{ojha2023towards}
uses features extracted by a large pre-trained model, CLIP, to detect synthetic images. After feature extraction, the dataset proposed in \cite{wang2020cnn} is used to design the classifier, testing various strategies, from nearest neighbor to linear probing.

\vspace{2mm}
\item{\bf NPR}~\cite{tan2023rethinking}
builds on the widespread use of up-sampling operations in generation models. proposing a simple way to represent up-sampling artifacts called Neighboring Pixel Relationships (NPR). NPR is computed from the difference between the original image and its interpolated version, which is then fed into a ResNet-50 trained on only four categories of the dataset proposed in \cite{wang2020cnn}.

\vspace{2mm}
\item{\bf AEROBLADE}~\cite{ricker2024aeroblade}
is a training-free approach that is based on the intuition that synthetic images are more accurately reconstructed by the autoencoder than real images. Then the discrimination is carried out by evaluating the reconstruction error of the autoencoder measured using the LPIPS distance between the image and its reconstructed version.

\end{enumerate}

\section{Datasets}
Tab.\ref{tab:dataset} lists all generators of synthetic images used in our experiments.
For each generator we provide the corresponding real data used in the test set,
the size of the generated images and
the number of images used in the experiments.
Overall, we are considering a large variety of synthetic generators of widely different characteristics for a total of 29k synthetic images and 6k real images.
These are a collection of datasets proposed in \cite{wang2020cnn, corvi2023detection, ojha2023towards, bammey2023synthbuster, kang2023scaling} and generated by ourselves.
It is worth noting that synthetic and real images are characterized by the same semantic content to avoid polarization. We also considered different reals from the same generator to understand the influence of the pristine class.

\setlength{\tabcolsep}{4pt}
\begin{table}[!t]
\caption{Details of datasets used in all experiments:
synthetic image generator, real images used in testing, number and size of images, source.}
\label{tab:dataset}
    \centering
\resizebox{0.70\textwidth}{!}{
\begin{tabular}{lcccc}
\toprule
\ru                    generator       & real data & \# images &            image size & source  \\ \midrule
\band                  GauGAN          & COCO      &      1000 & $256^{2}$             & \cite{wang2020cnn} \\
                       BigGAN          & ImageNet  &      1000 & $256^{2}$,$512^{2}$   & \cite{corvi2023detection} \\     
\band                  StarGAN         & FFHQ      &      1000 & $256^{2}$             & ours \\                                            
                                       & LSUN      &       500 & $256^{2}$             & \cite{corvi2023detection} \\                            
      \multirow{-2}{*}{StyleGAN2}      & FFHQ      &       500 & $256^{2}$,$1024^{2}$  & \cite{corvi2023detection}\\      
\band                                  & ImageNet  &       500 & $256^{2}$             & \cite{kang2023scaling} \\                                
\band \multirow{-2}{*}{GigaGAN}        & COCO      &       500 & $512^{2}$             & \cite{kang2023scaling} \\         
                       Diff.GAN        & LSUN      &      1000 & $256^{2}$             & ours \\                                           
\band                  GALIP           & COCO      &      1000 & $256^{2}$             & ours \\                                            
                       DALL·E          & LAION     &      1000 & $256^{2}$             & \cite{ojha2023towards} \\         
\band                  DDPM            & FFHQ      &      1000 & $256^{2}$             & ours \\                                 
                                       & LSUN      &       500 & $256^{2}$             & \cite{corvi2023detection} \\                            
      \multirow{-2}{*}{ADM}            & ImageNet  &       500 & $256^{2}$             & \cite{corvi2023detection} \\     
\band                                  & COCO      &      1000 & $256^{2}$             & \cite{corvi2023detection} \\                            
\band                                  & RAISE     &      1000 & $256^{2}$             & \cite{bammey2023synthbuster} \\                          
\band \multirow{-3}{*}{GLIDE}          & LAION     &      3000 & $256^{2}$             & \cite{ojha2023towards} \\
                       DiT             & ImageNet  &      1000 & $256^{2}$,$512^{2}$   & ours \\                                            
\band                                  & COCO      &      1000 & $256^{2}$             & ours \\                                                                   
\band \multirow{-2}{*}{Stable D. 1.4}  & RAISE     &      1000 & $512^{2}$             & \cite{bammey2023synthbuster} \\   
                                       & COCO      &      1000 & $256^{2}$-$768^{2}$   & ours \\                                                                   
      \multirow{-2}{*}{Stable D. 2}    & RAISE     &      1000 & $973^{2}$-$1024^{2}$  & \cite{bammey2023synthbuster} \\   
\band                                  & COCO      &      1000 & $1024^{2}$            & ours \\                                                                   
\band \multirow{-2}{*}{SDXL}           & RAISE     &      1000 & $973^{2}$-$1024^{2}$  & \cite{bammey2023synthbuster} \\   
                       Deep.-IF        & COCO      &      1000 & $1024^{2}$            & ours \\                                            
\band                                  & COCO      &      1000 & $1024^{2}$            & \cite{corvi2023detection} \\                            
\band \multirow{-2}{*}{DALL·E 2}       & RAISE     &      1000 & $1024^{2}$            & \cite{bammey2023synthbuster} \\    
                                       & COCO      &      1000 & $1024^{2}$            & ours \\                                                                  
      \multirow{-2}{*}{DALL·E 3}       & RAISE     &      1000 & $1024^{2}$-$1355^{2}$ & \cite{bammey2023synthbuster} \\   
\band                  Midjourney      & RAISE     &      1000 & $1024^{2}$-$1104^{2}$ & \cite{bammey2023synthbuster} \\   
                       Adobe Firefly   & RAISE     &      1000 & $2032^{2}$-$2048^{2}$ & \cite{bammey2023synthbuster} \\   
\bottomrule
\end{tabular}
}
\end{table}

\begin{figure}[!t]
    \centering
    \includegraphics[page=1,width=0.9\linewidth,clip,trim=0 0 0 30]{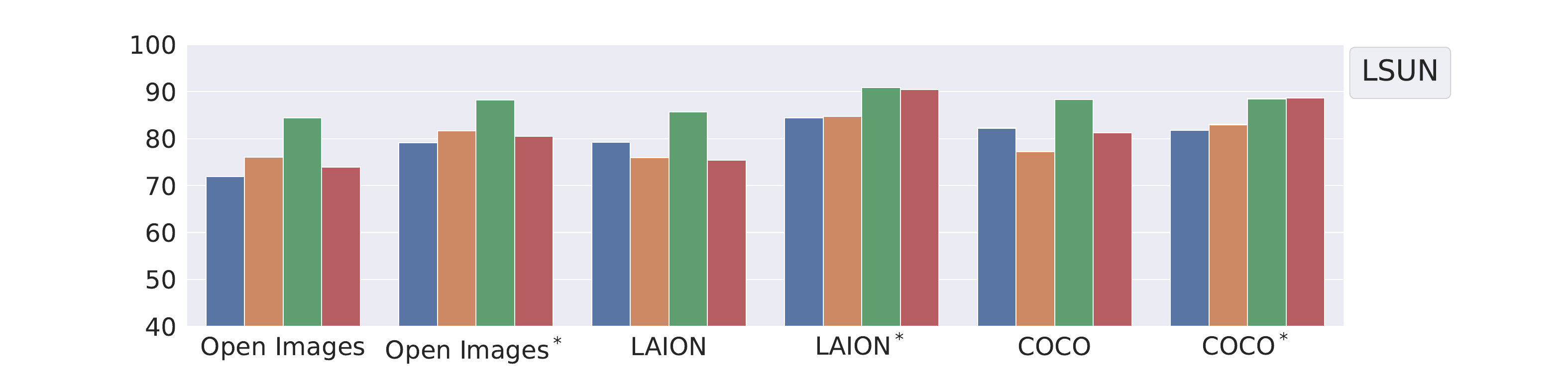} \\
    \includegraphics[page=2,width=0.9\linewidth,clip,trim=0 0 0 30]{figures/bars_db.pdf} \\
    \includegraphics[page=4,width=0.9\linewidth,clip,trim=0 0 0 30]{figures/bars_db.pdf} \\
    \includegraphics[page=5,width=0.9\linewidth,clip,trim=0 0 0 30]{figures/bars_db.pdf} \\
    \includegraphics[page=8,width=0.9\linewidth,clip,trim=0 0 0 30]{figures/bars_db.pdf} \\
    \includegraphics[page=7,width=0.5\linewidth,clip,trim=0 0 0 0]{figures/bars_db.pdf} 
    \caption{
Expanded ablation study, completing the results of figure 6 of main paper.
For each testing dataset (chart), 
and each training dataset (group of bars), 
we compute the AUC of the proposed method with four decision statistics (individual bars).
In each testing dataset, synthetic images are selected to match the corresponding real images. 
The final chart shows the average over all testing datasets.}
    \label{fig:other_ablation}
\end{figure}

\section{Additional ablation study}
\label{sec:other_ablation}

Here we describe a set of experiments designed to gain some insight into a few questions concerning training and testing:
\begin{description}
\item{Q1}: how much do training and testing datasets impact on performance?
\item{Q2}: how important is their alignment?
\item{Q3}: which are the best decision statistics?
\end{description}

To this end, we trained the lossless coder on three different datasets of real images: 
Open Images, LAION, and COCO.
In all cases, we considered two versions, with and without data augmentation.
Then we used six different testing datasets.
Each one includes the real images taken, in turn, from 
ImageNet, LAION, LSUN, FFHQ, COCO, and RAISE, with the associated fakes as outlined in Tab.\ref{tab:dataset}.
For example, the FFHQ testing dataset includes StarGAN, StyleGAN2, and DDPM synthetic images.
In all cases, we tested the four decision statistics defined in the main paper.

Results are reported in six charts, one for each testing dataset,
two of them already shown in Fig.\ref{fig:ablation} of the main paper 
and four more in Fig.\ref{fig:other_ablation}, 
followed by a final chart with average results.
From these results we see that, as expected, the quality of the training set has a significant impacts on performance.
By just considering the average (last chart) it is obvious that the coder trained on COCO with augmentation performs generally better than the others,  with some minor exceptions.
Likewise,
testing datasets differ appreciably from one another, for example ImageNet appears to be especially challenging under all conditions, followed by FFHQ.
Finally, the training-testing alignment does not seem to impact significantly on the performance.
For example, consider the charts corresponding to the LAION and COCO testing sets.
The coders trained on COCO appear to perform better than those trained on LAION not only on the COCO testing set, but also on the LAION testing set,
indicating that the quality of the training set prevails over the training-testing alignment.

We conclude with a note on the decision statistics.
Even though the overall best combination seems to be $|\Delta^{01}|$ with the coder trained on COCO with augmentation,
the $\Delta^{01}$ statistic (without modulus) ensures a very good performance uniformly over all trained encoders.
In any case results seem to vary very much from one dataset to the other.
For example $D^0$ works very well on RAISE and very badly on LAION.
This suggests that a much deeper analysis is necessary and better decision statistics are probably yet to be found.

\setlength{\tabcolsep}{4pt}
\begin{table}[!t]
    \centering
    \caption{Results in terms of AUC averaged for each real dataset. We have both JPEG compressed real images (COCO, IMAGENET, LSUN, LAION) and uncompressed images (FFHQ, RAISE).}
    \label{tab:comparative_dataset}
    \resizebox{1.0\textwidth}{!}
{\begin{tabular}{l !{\color{tabvline}\vrule} cccccccccc !{\color{tabvline}\vrule} cccc}
    \toprule
%%from python
\multicolumn{1}{c!{\color{tabvline}\vrule}}{\rotatebox{90}{Real data}} & \rotatebox{90}{Wang2020} & \rotatebox{90}{PatchFor.} & \rotatebox{90}{Liu2022} & \rotatebox{90}{Corvi2023} & \rotatebox{90}{LGrad} & \rotatebox{90}{DIRE} & \rotatebox{90}{DE-FAKE} & \rotatebox{90}{Ojha2023} & \rotatebox{90}{NPR} & \rotatebox{90}{AEROBLADE} & \rotatebox{90}{Ours $D^{(0)}$ } & \rotatebox{90}{Ours $\left|  D^{(0)} \right|$} & \rotatebox{90}{Ours $\Delta^{01}$} & \rotatebox{90}{Ours $\left| \Delta^{01} \right|$} \\
\midrule 
\band      COCO &    77.3  &    74.8  &    93.9  &    94.8  &    73.9  & \b{99.7} &    85.8  &    92.2  &    91.5  &    82.7  &    86.5  &    92.5  &    91.3  &    92.2  \\           
       IMAGENET &    71.5  &    77.9  &    93.6  &    78.8  &    70.6  & \b{99.7} &    69.9  &    89.3  &    81.5  &    69.9  &    85.5  &    82.9  &    89.6  &    85.9  \\
\band      LSUN &    85.2  &    81.3  &    89.9  &    88.6  &    93.5  &    53.4  &    40.4  &    94.1  & \b{98.0} &    42.3  &    81.8  &    83.0  &    88.5  &    89.0  \\
          LAION &    69.5  &    72.9  &    92.9  &    92.6  &    92.8  & \b{99.9} &    58.1  &    96.3  & \b{99.7} &    46.4  &    76.8  &    90.0  &    96.0  &    91.9  \\ \midrule
\band      FFHQ &    73.7  & \b{94.5} &    74.2  &    84.8  &    55.5  &    39.7  &    48.0  &    89.5  &    81.8  &    75.3  &    73.2  &    71.0  &    78.3  &    89.4  \\
          RAISE &    46.9  &    58.2  &    39.9  & \b{89.3} &    49.1  &    45.5  &    83.5  &    78.8  &    53.0  &    73.1  &    84.2  &    86.5  &    87.5  & \b{89.3} \\
%%end python
\bottomrule
    \end{tabular}}
\end{table}

\section{Effect of the dataset JPEG bias}

Recent papers \cite{grommelt2024fake,ricker2024aeroblade} highlighted the presence of a bias in some datasets used in the field where all real images are JPEG compressed, while the generated images are stored in lossless format.
Of course, if a detector is trained and tested on datasets affected by this bias, its results may be significantly altered.   
Our method is not affected by such distortion since it relies on a lossless encoder that provides an intrinsic model of real images which does not fit well to synthetic images.
To confirm this fact, in Tab.\ref{tab:comparative_dataset} we report the same AUC results already shown in the main paper, but now averaged over each real dataset.
At the top we have datasets with compressed images and at the bottom datasets with uncompressed images.
Therefore, at the bottom we have a situation where both real and synthetic images are uncompressed and no JPEG-related distortion can affect the results.
For our method there is no significant difference in performance between the top and bottom of the table.
On the contrary, some SoTA methods show a more controversial behavior.

\section{Additional comparison}

\begin{figure}[!t]
    \centering
    \includegraphics[page=1, trim=25 5 0 0, clip,width=0.24\linewidth]{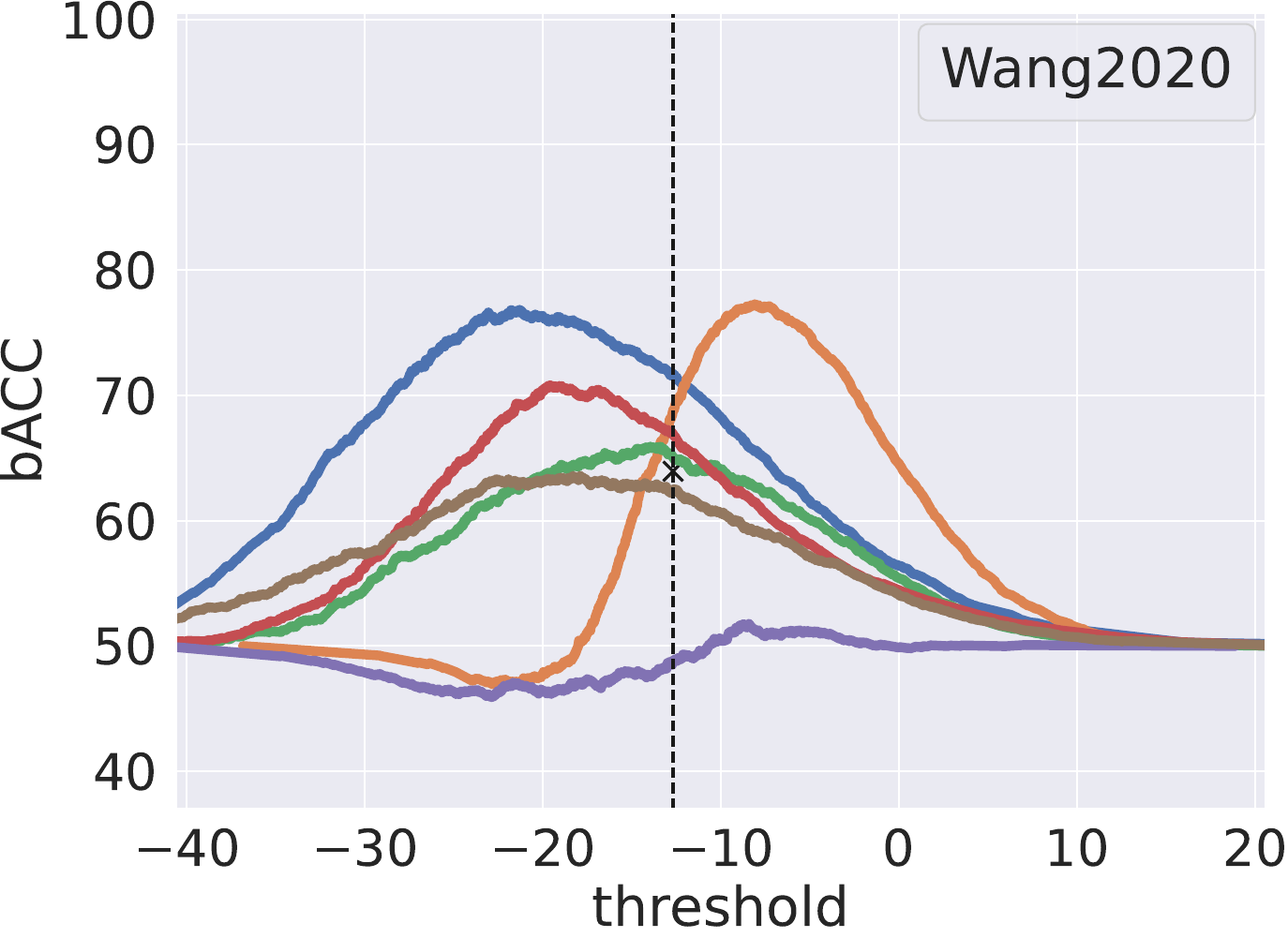} 
    \includegraphics[page=2, trim=25 5 0 0, clip,width=0.24\linewidth]{figures/graph_ths.pdf} 
    \includegraphics[page=3, trim=25 5 0 0, clip,width=0.24\linewidth]{figures/graph_ths.pdf} 
    \includegraphics[page=5, trim=25 5 0 0, clip,width=0.24\linewidth]{figures/graph_ths.pdf} \\
    \includegraphics[page=6, trim=25 0 0 0, clip,width=0.24\linewidth]{figures/graph_ths.pdf} 
    \includegraphics[page=7, trim=25 0 0 0, clip,width=0.24\linewidth]{figures/graph_ths.pdf} 
    \includegraphics[page=10,trim=25 0 0 0, clip,width=0.24\linewidth]{figures/graph_ths.pdf}
    \includegraphics[page=11,trim=25 0 0 0, clip,width=0.24\linewidth]{figures/graph_ths.pdf} \\
    \includegraphics[page=12,width=0.7\linewidth]{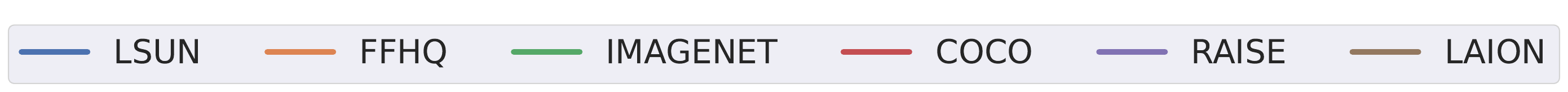}
    \caption{
    Balanced accuracy as a function of the detection threshold for methods not analyzed in the main paper. For each dataset of real images, we average accuracy over all associated synthetic generators. The dotted vertical line indicates the global optimal threshold and the $\times$ symbol the corresponding accuracy. Again, only for the proposed method all peaks are very close, indicating the presence of a single threshold.}
    \label{fig:graph_ths_all}
\end{figure}

In Fig.\ref{fig:graph_ths} of the main paper, we report results in terms of balanced accuracy only for the proposed method and the best three competitors. For completeness, in Fig.\ref{fig:graph_ths_all} we show the results of the other seven competitors compared again with those of the proposal. Like for the other competitors, the best threshold varies considerably depending on the dataset and hence no single threshold can provide good results in all cases.

\end{appendix}

\bibliographystyle{splncs04}
\bibliography{egbib}

\begin{thebibliography}{10}
\providecommand{\url}[1]{\texttt{#1}}
\providecommand{\urlprefix}{URL }
\providecommand{\doi}[1]{https://doi.org/#1}

\bibitem{albright2019source}
Albright, M., McCloskey, S.: {Source Generator Attribution via Inversion}. In: CVPR Workshop. pp. 96--103 (2019)

\bibitem{amoroso2023parents}
Amoroso, R., Morelli, D., Cornia, M., Baraldi, L., Del~Bimbo, A., Cucchiara, R.: {Parents and Children: Distinguishing Multimodal DeepFakes from Natural Images}. ACM Trans. Multimedia Comput. Commun. Appl.  (2024)

\bibitem{bammey2023synthbuster}
Bammey, Q.: {Synthbuster: Towards Detection of Diffusion Model Generated Images}. IEEE Open Journal of Signal Processing  (2023)

\bibitem{bohacek2023geometric}
Boh\'a\v{c}ek, M., Farid, H.: {A geometric and photometric exploration of GAN and Diffusion synthesized faces}. In: CVPR Workshop. pp. 874–--883 (2023)

\bibitem{brock2018large}
Brock, A., Donahue, J., Simonyan, K.: {Large Scale GAN Training for High Fidelity Natural Image Synthesis}. In: ICLR (2018)

\bibitem{cao2020lossless}
Cao, S., Wu, C.Y., Kr{\"{a}}henb{\"{u}}hl, P.: {Lossless Image Compression through Super-Resolution}. arXiv preprint arXiv:2004.02872v1  (2020)

\bibitem{chai2020what}
Chai, L., Bau, D., Lim, S.N., Isola, P.: {What Makes Fake Images Detectable? Understanding Properties that Generalize}. In: ECCV. pp. 103--120 (2020)

\bibitem{choi2018stargan}
Choi, Y., Choi, M., Kim, M., Ha, J.W., Kim, S., Choo, J.: {StarGAN}: Unified generative adversarial networks for multi-domain image-to-image translation. In: CVPR. pp. 8789--8797 (2018)

\bibitem{corvi2023intriguing}
Corvi, R., Cozzolino, D., Poggi, G., Nagano, K., Verdoliva, L.: Intriguing properties of synthetic images: from generative adversarial networks to diffusion models. In: CVPR Workshop. pp. 973--982 (2023)

\bibitem{corvi2023detection}
Corvi, R., Cozzolino, D., Zingarini, G., Poggi, G., Nagano, K., Verdoliva, L.: On the detection of synthetic images generated by diffusion models. In: ICASSP. pp.~1--5 (2023)

\bibitem{cozzolino2023raising}
Cozzolino, D., Poggi, G., Corvi, R., Nie{\ss}ner, M., Verdoliva, L.: {Raising the Bar of AI-generated Image Detection with CLIP}. In: CVPR Workshop. pp. 4356--4366 (2024)

\bibitem{cozzolino2018forensictransfer}
Cozzolino, D., Thies, J., R{\"o}ssler, A., Riess, C., Nie{\ss}ner, M., Verdoliva, L.: {Forensictransfer: Weakly-supervised domain adaptation for forgery detection}. arXiv preprint arXiv:1812.02510  (2018)

\bibitem{nguyen2015raise}
Dang-Nguyen, D.T., Pasquini, C., Conotter, V., Boato, G.: {RAISE: A Raw Images Dataset for Digital Image Forensics}. In: {ACM MMSys}. p. 219–224 (2015)

\bibitem{Dayma_DALLE_Mini_2021}
Dayma, B., Patil, S., Cuenca, P., Saifullah, K., Abraham, T., Lê~Khàc, P., Melas, L., Ghosh, R.: {DALL-E Mini} (2021). \doi{10.5281/zenodo.5146400}, \url{https://github.com/borisdayma/dalle-mini}

\bibitem{deng2009imagenet}
Deng, J., Dong, W., Socher, R., Li, L.J., Li, K., Fei-Fei, L.: {ImageNet: A large-scale hierarchical image database}. In: CVPR. pp. 248--255 (2009)

\bibitem{dhariwal2021diffusion}
Dhariwal, P., Nichol, A.: {Diffusion models beat GANs on image synthesis}. NeurIPS  \textbf{34},  8780--8794 (2021)

\bibitem{du2020towards}
Du, M., Pentyala, S., Li, Y., Hu, X.: {Towards Generalizable Deepfake Detection with Locality-Aware AutoEncoder}. In: CIKM. pp. 325--–334 (2020)

\bibitem{durall2020watch}
Durall, R., Keuper, M., Keuper, J.: {Watch Your Up-Convolution: CNN Based Generative Deep Neural Networks Are Failing to Reproduce Spectral Distributions}. In: CVPR. pp. 7890--7899 (2020)

\bibitem{epstein2023online}
Epstein, D.C., Jain, I., Wang, O., Zhang, R.: {Online Detection of AI-Generated Images}. In: ICCV Workshop. pp. 382--392 (2023)

\bibitem{epstein2023art}
Epstein, Z., Hertzmann, A., Herman, L., Mahari, R., Frank, M.R., Groh, M., Schroeder, H., Akten, A.S.M., Fjeld, J., Farid, H., Leach, N., Pentland, A.S., Russakovsky, O.: {Art and the science of generative AI: A deeper dive}. arXiv preprint arXiv:2306.04141  (2023)

\bibitem{farid2022lighting}
Farid, H.: {Lighting (in) consistency of paint by text}. arXiv preprint arXiv:2207.13744  (2022)

\bibitem{farid2022perspective}
Farid, H.: {Perspective (in) consistency of paint by text}. arXiv preprint arXiv:2206.14617  (2022)

\bibitem{firefly}
Firefly, A.: \url{https://www.adobe.com/sensei/generative-ai/firefly.html} (2023)

\bibitem{frank2020leveraging}
Frank, J., Eisenhofer, T., Sch{\"o}nherr, L., Fischer, A., Kolossa, D., Holz, T.: {Leveraging Frequency Analysis for Deep Fake Image Recognition}. In: ICML. pp. 3247--3258 (2020)

\bibitem{gehrmann2019gltr}
Gehrmann, S., Strobelt, H., Rush, A.M.: {GLTR: Statistical detection and visualization of generated text}. In: {57th Annual Meeting of the Association for Computational Linguistics: System Demonstrations}. pp. 111--116 (2019)

\bibitem{ghosal2023towards}
Ghosal, S.S., Chakraborty, S., Geiping, J., Huang, F., Manocha, D., Bedi, A.S.: {Towards possibilities \& impossibilities of AI-generated text detection: A survey}. arXiv preprint arXiv:2310.15264  (2023)

\bibitem{gragnaniello2021GAN}
Gragnaniello, D., Cozzolino, D., Marra, F., Poggi, G., Verdoliva, L.: {Are GAN generated images easy to detect? A critical analysis of the state-of-the-art}. In: ICME. pp.~1--6 (2021)

\bibitem{grommelt2024fake}
Grommelt, P., Weiss, L., Pfreundt, F.J., Keuper, J.: {Fake or JPEG? Revealing Common Biases in Generated Image Detection Datasets}. arXiv preprint arXiv:2403.17608  (2024)

\bibitem{hans2024spotting}
Hans, A., Schwarzschild, A., Cherepanova, V., Kazemi, H., Saha, A., Goldblum, M., Geiping, J., Goldstein, T.: {Spotting LLMs With Binoculars: Zero-Shot Detection of Machine-Generated Text}. In: ICML (2024)

\bibitem{he2024rigid}
He, Z., Chen, P.Y., Ho, T.Y.: {RIGID: A Training-free and Model-Agnostic Framework for Robust AI-Generated Image Detection}. arXiv preprint arXiv:2405.20112  (2024)

\bibitem{heikkila2022this}
Heikkil{\"a}, M.: {This artist is dominating AI-generated art. and he’s not happy about it}. MIT Technology Review  (2022)

\bibitem{ho2020denoising}
Ho, J., Jain, A., Abbeel, P.: Denoising diffusion probabilistic models. NeurIPS  \textbf{33},  6840--6851 (2020)

\bibitem{jeon2020tgd}
Jeon, H., Bang, Y.O., Kim, J., Woo, S.: {T-GD: Transferable {GAN}-generated Images Detection Framework}. In: ICML. vol.~119, pp. 4746--4761 (2020)

\bibitem{jeong2022FingerprintNet}
Jeong, Y., Kim, D., Ro, Y., Kim, P., Choi, J.: {FingerprintNet: Synthesized Fingerprints for Generated Image Detection}. In: ECCV. pp. 76--94 (2022)

\bibitem{kang2023scaling}
Kang, M., Zhu, J.Y., Zhang, R., Park, J., Shechtman, E., Paris, S., Park, T.: Scaling up gans for text-to-image synthesis. In: CVPR. pp. 10124--10134 (2023)

\bibitem{karras2018progressive}
Karras, T., Aila, T., Laine, S., Lehtinen, J.: {Progressive Growing of GANs for Improved Quality, Stability, and Variation}. In: ICLR (2018)

\bibitem{karras2019stylegan}
Karras, T., Laine, S., Aila, T.: A style-based generator architecture for generative adversarial networks. In: CVPR. pp. 4401--4410 (2019)

\bibitem{karras2020stylegan2}
Karras, T., Laine, S., Aittala, M., Hellsten, J., Lehtinen, J., Aila, T.: {Analyzing and improving the image quality of StyleGAN}. In: CVPR. pp. 8110--8119 (2020)

\bibitem{deepfloydif}
Konstantinov, M., Shonenkov, A., Bakshandaeva, D., Schuhmann, C., Ivanova, K., Klokova, N.: \url{https://www.deepfloyd.ai/deepfloyd-if} (2023)

\bibitem{krasin2017openimages}
Krasin, I., Duerig, T., Alldrin, N., Ferrari, V., Abu-El-Haija, S., Kuznetsova, A., Rom, H., Uijlings, J., Popov, S., Veit, A., {et al.}: {OpenImages: A public dataset for large-scale multi-label and multi-class image classification.} Dataset available from \url{https://github.com/openimages}  (2017)

\bibitem{lin2024detecting}
Lin, L., Gupta, N., Zhang, Y., Ren, H., Liu, C.H., Ding, F., Wang, X., Li, X., Verdoliva, L., Hu, S.: Detecting multimedia generated by large ai models: A survey. arXiv preprint arXiv:2204.06125  (2024)

\bibitem{lin2014microsoft}
Lin, T.Y., Maire, M., Belongie, S., Hays, J., Perona, P., Ramanan, D., Doll{\'a}r, P., Zitnick, C.L.: {Microsoft COCO: Common objects in context}. In: ECCV. pp. 740--755 (2014)

\bibitem{liu2022detecting}
Liu, B., Yang, F., Bi, X., Xiao, B., Li, W., Gao, X.: Detecting generated images by real images. In: ECCV. pp. 95--110 (2022)

\bibitem{liu2023forgery}
Liu, H., Tan, Z., Tan, C., Wei, Y., Wang, J., Zhao, Y.: {Forgery-aware Adaptive Transformer for Generalizable Synthetic Image Detection}. In: CVPR. pp. 10770--10780 (2024)

\bibitem{mahajan2021pixelpyramids}
Mahajan, S., Roth, S.: {PixelPyramids: Exact Inference Models from Lossless Image Pyramids}. In: ICCV. pp. 6639--6648 (2021)

\bibitem{mandelli2022detecting}
Mandelli, S., Bonettini, N., Bestagini, P., Tubaro, S.: {Detecting GAN-generated Images by Orthogonal Training of Multiple CNNs}. In: ICIP. pp. 3091--3095 (2022)

\bibitem{marra2019incremental}
Marra, F., Saltori, C., Boato, G., Verdoliva, L.: {Incremental learning for the detection and classification of GAN-generated images}. In: WIFS. pp.~1--6 (2019)

\bibitem{midjourney}
Midjourney: \url{https://www.midjourney.com/home} (2023)

\bibitem{mitchell2023detectGLPT}
Mitchell, E., Lee, Y., Khazatsky, A., Manning, C.D., Finn, C.: {DetectGPT: Zero-Shot Machine-Generated Text Detection using Probability Curvature}. In: ICML. pp. 24950--24962 (2023)

\bibitem{nichol2021glide}
Nichol, A.Q., Dhariwal, P., Ramesh, A., Shyam, P., Mishkin, P., Mcgrew, B., Sutskever, I., Chen, M.: {GLIDE: Towards Photorealistic Image Generation and Editing with Text-Guided Diff. Models}. In: ICML. pp. 16784--16804 (2022)

\bibitem{ojha2023towards}
Ojha, U., Li, Y., Lee, Y.J.: Towards universal fake image detectors that generalize across generative models. In: CVPR. pp. 24480--24489 (2023)

\bibitem{dalle3}
OpenAI: \url{https://openai.com/dall-e-3} (2023)

\bibitem{park2019semantic}
Park, T., Liu, M.Y., Wang, T.C., Zhu, J.Y.: Semantic image synthesis with spatially-adaptive normalization. In: CVPR. pp. 2337--2346 (2019)

\bibitem{peebles2022scalable}
Peebles, W., Xie, S.: Scalable diffusion models with transformers. In: ICCV. pp. 4195--4205 (2023)

\bibitem{podell2023sdxl}
Podell, D., English, Z., Lacey, K., Blattmann, A., Dockhorn, T., M{\"u}ller, J., Penna, J., Rombach, R.: {SDXL}: Improving latent diffusion models for high-resolution image synthesis. In: ICLR (2024)

\bibitem{radford2021learning}
Radford, A., Kim, J.W., Hallacy, C., Ramesh, A., Goh, G., Agarwal, S., Sastry, G., Askell, A., Mishkin, P., Clark, J., et~al.: Learning transferable visual models from natural language supervision. In: ICML. pp. 8748--8763 (2021)

\bibitem{ramesh2022hierarchical}
Ramesh, A., Dhariwal, P., Nichol, A., Chu, C., Chen, M.: {Hierarchical Text-Conditional Image Generation with CLIP Latents}. arXiv preprint arXiv:2204.06125  (2022)

\bibitem{reed2017parallel}
Reed, S.E., van~den Oord, A., Kalchbrenner, N., Colmenarejo, S.G., Wang, Z., Chen, Y., Belov, D., de~Freitas, N.: {Parallel multiscale autoregressive density estimation}. In: ICML. pp. 2912--2921 (2017)

\bibitem{ricker2022towards}
Ricker, J., Damm, S., Holz, T., Fischer, A.: {Towards the detection of diffusion model deepfakes}. In: {VISAPP}. pp. 446--457 (2024)

\bibitem{ricker2024aeroblade}
Ricker, J., Lukovnikov, D., Fischer, A.: {AEROBLADE: Training-Free Detection of Latent Diffusion Images Using Autoencoder Reconstruction Error}. In: CVPR. pp. 9130--9140 (2024)

\bibitem{rombach2022high}
Rombach, R., Blattmann, A., Lorenz, D., Esser, P., Ommer, B.: High-resolution image synthesis with latent diffusion models. In: CVPR. pp. 10684--10695 (2022)

\bibitem{stablediffusion1}
Rombach, R., Blattmann, A., Lorenz, D., Esser, P., Ommer, B.: \url{https://github.com/CompVis/stable-diffusion} (2022)

\bibitem{stablediffusion2}
Rombach, R., Blattmann, A., Lorenz, D., Esser, P., Ommer, B.: \url{https://github.com/Stability-AI/stablediffusion} (2022)

\bibitem{roessler2019ff++}
R{\"{o}}ssler, A., Cozzolino, D., Verdoliva, L., Riess, C., Thies, J., Nie{\ss}ner, M.: {Faceforensics++: Learning to detect manipulated facial images}. In: ICCV. pp. 1--11 (2019)

\bibitem{sarkar2023shadows}
Sarkar, A., Mai, H., Mahapatra, A., Lazebnik, S., Forsyth, D.A., Bhattad, A.: {Shadows Don't Lie and Lines Can't Bend! Generative Models don't know Projective Geometry... for now}. In: CVPR. pp. 28140--28149 (2024)

\bibitem{schuhmann2021laion}
Schuhmann, C., Kaczmarczyk, R., Komatsuzaki, A., Katta, A., Vencu, R., Beaumont, R., Jitsev, J., Coombes, T., Mullis, C.: {LAION-400M: Open Dataset of CLIP-Filtered 400 Million Image-Text Pairs}. In: NeurIPS (2021)

\bibitem{sha2022fake}
Sha, Z., Li, Z., Yu, N., Zhang, Y.: {DE-FAKE: Detection and Attribution of Fake Images Generated by Text-to-Image Generation Models}. In: ACM SIGSAC. pp. 3418--3432 (2023)

\bibitem{sinitsa2023deep}
Sinitsa, S., Fried, O.: {Deep Image Fingerprint: Towards Low Budget Synthetic Image Detection and Model Lineage Analysis}. In: WACV. pp. 4067--4076 (2024)

\bibitem{solaiman2019release}
Solaiman, I., Brundage, M., Clark, J., Askell, A., Herbert-Voss, A., Wu, J., Radford, A., Krueger, G., Kim, J.W., Kreps, S., et~al.: {Release Strategies and the Social Impacts of Language Models}. arXiv preprint arXiv:1908.09203  (2019)

\bibitem{su2023detectllm}
Su, J., Zhuo, T.Y., Wang, D., Nakov, P.: {DetectLLM: Leveraging Log Rank Information for Zero-Shot Detection of Machine-Generated Text}. In: Conference on Empirical Methods in Natural Language Processing (2023)

\bibitem{tan2023rethinking}
Tan, C., Zhao, Y., Wei, S., Gu, G., Liu, P., Wei, Y.: {Rethinking the Up-Sampling Operations in CNN-based Generative Network for Generalizable Deepfake Detection}. In: CVPR. pp. 28130--28139 (2024)

\bibitem{tan2023learning}
Tan, C., Zhao, Y., Wei, S., Gu, G., Wei, Y.: {Learning on Gradients: Generalized Artifacts Representation for GAN-Generated Images Detection}. In: CVPR. pp. 12105--12114 (2023)

\bibitem{tao2023galip}
Tao, M., Bao, B.K., Tang, H., Xu, C.: Galip: Generative adversarial clips for text-to-image synthesis. In: CVPR. pp. 14214--14223 (2023)

\bibitem{wang2020cnn}
Wang, S.Y., Wang, O., Zhang, R., Owens, A., Efros, A.A.: {CNN-generated images are surprisingly easy to spot... for now}. In: CVPR. pp. 8692--8701 (2020)

\bibitem{wang2023dire}
Wang, Z., Bao, J., Zhou, W., Wang, W., Hu, H., Chen, H., Li, H.: {DIRE for Diffusion-Generated Image Detection}. ICCV pp. 22445--22455 (2023)

\bibitem{wang2023diffusion}
Wang, Z., Zheng, H., He, P., Chen, W., Zhou, M.: {Diffusion-GAN: Training GANs with Diffusion}. In: ICLR (2023)

\bibitem{yu2015lsun}
Yu, F., Seff, A., Zhang, Y., Song, S., Funkhouser, T., Xiao, J.: {LSUN: Construction of a large-scale image dataset using deep learning with humans in the loop}. arXiv preprint arXiv:1506.03365  (2015)

\bibitem{zhang2019detecting}
Zhang, X., Karaman, S., Chang, S.F.: {Detecting and Simulating Artifacts in GAN Fake Images}. In: WIFS. pp.~1--6 (2019)

\bibitem{zhong2023rich}
Zhong, N., Xu, Y., Qian, Z., Zhang, X.: {Rich and Poor Texture Contrast: A Simple yet Effective Approach for AI-generated Image Detection}. arXiv preprint arXiv:2311.12397v1  (2023)

\end{thebibliography}

\end{document}